\documentclass[10pt,twocolumn,letterpaper]{article}

\usepackage[pagenumbers]{cvpr} 

\definecolor{cvprblue}{rgb}{0.21,0.49,0.74}
\usepackage[pagebackref,breaklinks,colorlinks,allcolors=cvprblue]{hyperref}

\usepackage{multirow}
\usepackage{bbding}
\usepackage{algorithm}
\usepackage{algorithmic}

\title{Siamese Machine Unlearning with Knowledge Vaporization and Concentration}

\author{Songjie Xie, Hengtao He, Shenghui Song, Jun Zhang, and Khaled~B.~Letaief\\
Dept. of Electronic and Computer Engineering, The Hong Kong University of Science and Technology\\
{\tt\small sxieat@connect.ust.hk, \{eehthe, eeshsong, eejzhang, eekhaled\}@ust.hk}
}

\begin{document}
\maketitle
\newcommand{\x}{\mathbf{x}}
\newcommand{\z}{\mathbf{z}}
\newcommand{\y}{\mathbf{y}}
\newcommand{\m}{\mathbf{m}}
\newcommand{\hz}{\hat{\mathbf{z}}}
\newcommand{\hy}{\hat{\mathbf{y}}}
\newcommand{\hZ}{\hat{Z}}
\newcommand{\hY}{\hat{Y}}
\newcommand{\bE}{\mathbb{E}}
\newcommand{\bR}{\mathbb{R}}

\newcommand{\D}{\mathcal{D}}

\newcommand{\bPhi}{\boldsymbol{\phi}}
\newcommand{\bTheta}{\boldsymbol{\theta}}
\newcommand{\bSigma}{\boldsymbol{\sigma}}
\newcommand{\bomega}{\boldsymbol{\omega}}

\renewcommand{\algorithmicrequire}{ \textbf{Input:}} 
\renewcommand{\algorithmicensure}{ \textbf{Output:}} 
\newtheorem{definition}{Definition} 
\newtheorem{remark}{Remark}
\begin{abstract}
In response to the practical demands of the ``right to be forgotten" and the removal of undesired data, machine unlearning emerges as an essential technique to remove the learned knowledge of a fraction of data points from trained models. However, existing methods suffer from limitations such as insufficient methodological support, high computational complexity, and significant memory demands. In this work, we propose the concepts of knowledge vaporization and concentration to selectively erase learned knowledge from specific data points while maintaining representations for the remaining data. Utilizing the Siamese networks, we exemplify the proposed concepts and develop an efficient method for machine unlearning. Our proposed Siamese unlearning method does not require additional memory overhead and full access to the remaining dataset. Extensive experiments conducted across multiple unlearning scenarios showcase the superiority of Siamese unlearning over baseline methods, illustrating its ability to effectively remove knowledge from forgetting data, enhance model utility on remaining data, and reduce susceptibility to membership inference attacks.
\end{abstract}
\section{Introduction}
\label{sec:1_intro}
Modern machine learning (ML) models play a pivotal role across diverse domains, serving as the fundamental infrastructure for numerous human-centric applications. However, their substantial reliance on extensive datasets during training gives rise to significant privacy concerns. In response to these pressing privacy concerns, the contemporary and forthcoming governmental regulations, such as the General Data Protection and Regulation (GDPR)~\cite{voigt2017eu} and the AI Act~\cite{veale2021demystifying}, mandate deletion-upon-request of user data and include the right to be forgotten. To adhere to these emerging obligations, there is a surge of interest in \emph{machine unlearning}, which aims to erase the influence of specific data samples from a trained model.

The most direct approach to achieve unlearning is to retrain ML models from scratch without using the data to be unlearned. Although this naive method yields the ground-truth unlearning strategy, its prohibitive computational demand hinders its application in real-world scenarios. To this end, a body of works have been proposed to expedite the retraining process, called \emph{exact unlearning}. These methods, such as retaining batch gradients from the initial training phase~\cite{graves2021amnesiac} and partitioning the model into sub-models for aggregation~\cite{bourtoule2021machine}, aim to accelerate the retraining-based unlearning. Nevertheless, existing exact unlearning methods suffer from significant computational complexity, increased memory requirements, and potential deterioration in model performance~\cite{nguyen2022survey}. To address these issues, recent studies have shifted focus towards \emph{approximate unlearning}, which aims to adjust the weights of the trained model to approximate the model obtained through retraining.

Approximate unlearning emerges as a more practical approach that reduces the influence of forgetting data through efficient updates for the model parameters. Recent works in approximate unlearning leverage a diverse range of methodologies based on distinct conceptualizations and viewpoints on approximation. For instance, techniques utilizing the influence function~\cite{koh2017understanding} and the Fisher information matrix (FIM)~\cite{golatkar2020eternal, martens2020new} are adopted to identify and mitigate the impact of forgotten data on model parameters. However, these methods have high computational complexity owing to the high dimensionality of the parameter space. Knowledge distillation-based methods~\cite{chundawat2023can, kurmanji2024towards}  adopt teacher-student frameworks to selectively scrub knowledge learned from forgetting data, but there are still memory consumption concerns as they require maintaining one or multiple teacher models. Boundary unlearning~\cite{chen2023boundary} shifts focus from the parameter space to the decision space, facilitating the rapid elimination of the target class. Nonetheless, this technique is developed to erase the whole class, and cannot be applied in instance-wise unlearning scenarios. 

Note that approximate unlearning fundamentally strives to approximate the retrained model. To find an effective way to achieve this approximation, we visualize the output logits of retrained models across augmentations of both forgetting and remaining data.
\begin{figure}[t]
\begin{subfigure}{\linewidth}
  \centering
  \includegraphics[width=.99\linewidth]{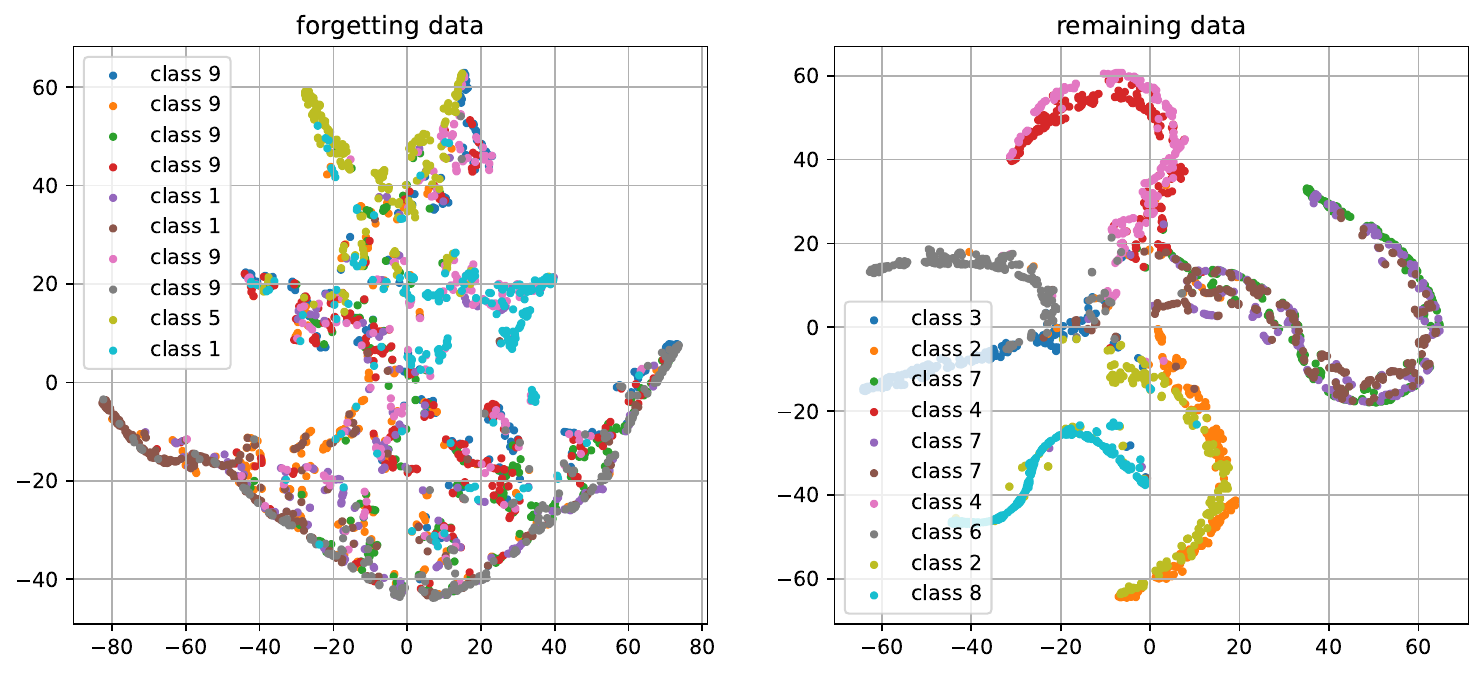}
  \caption{Forgetting full-classes of \texttt{automobile}, \texttt{cat}, and \texttt{truck}}
  \label{subfig:intro_fig_1}
\end{subfigure}%

\begin{subfigure}{\linewidth}
  \centering
  \includegraphics[width=.99\linewidth]{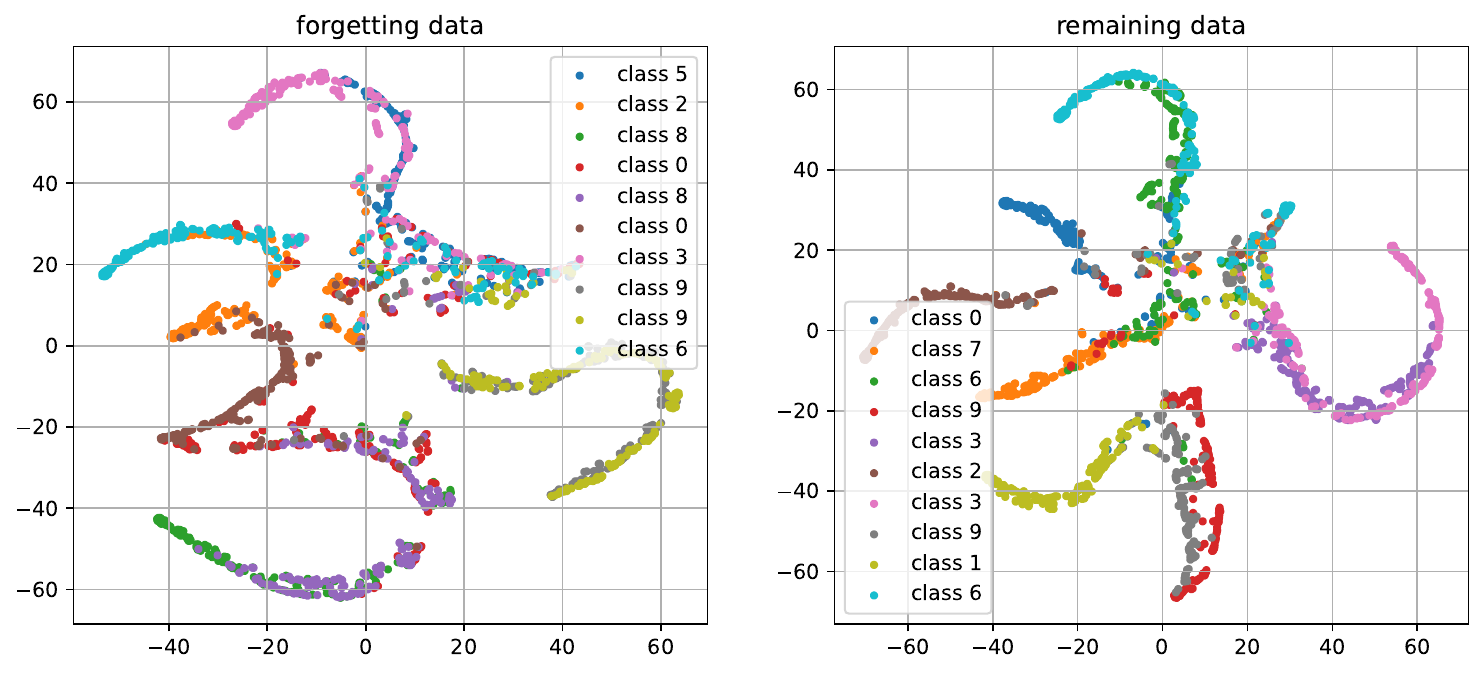}
  \caption{Forgetting 10\% random data samples }
  \label{subfig:intro_fig_2}
\end{subfigure}
\caption{$t$-SNE visualization of logit outputs from retrained models in two scenarios: (a) full-class forgetting and (b) random forgetting. Dots in different colors represent augmented views of different data points. The left panel shows the visualization for the augmented views of forgetting data, while the right panel displays those of remaining data. In both scenarios, it can be observed that the augmented views of forgetting samples exhibit greater dispersion compared to those of remaining samples, with no distinct clusters.} 
\vspace{-10pt}
\label{fig:intro_fig}
\end{figure}
As shown in Figure~\ref{fig:intro_fig}, we observe that augmented views of forgetting data are dispersed over the embedding space, even for the same data point. We also calculate the Kullback–Leibler divergence for the augmented views of the same data points, revealing that forgetting data exhibits higher values than the remaining data (see Appendix A for more details). This observation aligns with our primary goal and the intrinsic nature of the unlearning process. From the privacy perspective, the output of unlearned models should be sensitive to data augmentation and display increased uncertainty to minimize information leakage associated with forgetting data. Furthermore, from a learning standpoint, the failure of neural networks to converge to a stable representation across augmentations of forgetting data implies the absence of retrievable knowledge from these instances. Consequently, data augmentation and output logits play crucial roles in determining whether the learned knowledge of a data sample has been effectively erased from the models. This principle has been taken oppositely by contrastive learning for assessing whether a self-supervised model is truly learning from data~\cite{tian2020makes, chen2020simple}.

Based on the ideas outlined above, we introduce the concept of \emph{knowledge vaporization}, which is to effectively eliminate the learned knowledge from specific data points by dispersing the corresponding logits randomly across the space. Furthermore, to prevent knowledge vaporization from compromising the utility of the model, we concurrently propose a contrasting approach named \emph{knowledge concentration}, which aims to preserve the learned representations for the remaining data. The synergy of knowledge vaporization and concentration facilitates efficient instance-specific unlearning. Additionally, to avoid the memory overhead of teacher-student frameworks, we veer away from conventional methods and instead opt for Siamese networks to exemplify the processes of knowledge vaporization and concentration. These processes are performed with adaptively permuted labels for forgetting data and a few samples of remaining data, enhancing the effectiveness of the unlearning mechanism.

We summarize our major contributions as follows:
\begin{itemize}
    \item We propose the concepts of knowledge vaporization and concentration, which collaboratively target the selective removal of learned knowledge from a trained model.
    \item Based on the proposed knowledge vaporization and concentration, we develop an efficient unlearning method utilizing Siamese networks without adding additional memory overhead to the original model.
    \item  We conduct extensive experiments on multiple unlearning scenarios to evaluate the proposed unlearning methods. The results show that Siamese unlearning outperforms baseline methods by effectively erasing the knowledge from forgetting data, enhancing model utility on retained data, while demonstrating reduced susceptibility to membership inference attacks.
\end{itemize}
\section{Related Work}
\label{sec:2_related}
\paragraph{Exact unlearning.} Exact unlearning, also called retrain acceleration, aims to eliminate the influence of targeted data points from the model through rapid retraining. Many works have been proposed to accelerate the retraining of conventional ML algorithms such as statistical query learning~\cite{cao2015towards} and $k$-means~\cite{ginart2019making}. For the neural network-based models, \citet{ullah2021machine} introduced exact unlearning by storing historical model parameters during the training phase of empirical risk minimization. Another popular approach, SISA~\cite{bourtoule2021machine}, was proposed to divide the training data into several segments and train sub-models on each segment. It is more efficient as it only needs to retrain the affected segment when unlearning requests arise. Additionally, \citet{chen2022graph} developed an exact unlearning for graph neural networks based on the SISA framework. While providing strong guarantees of removal, exact unlearning still imposes significant computational demands, increased memory requirements, and potential performance degradation.

\paragraph{Approximate unlearning.} Approximate unlearning~\cite{golatkar2020eternal, graves2021amnesiac, chen2023boundary, lin2023erm, kurmanji2024towards} is another broader category of methods that avoids the need for retraining by reducing the influence of forgetting data through efficient parameter updates. As a straightforward attempt, \citet{golatkar2020eternal} proposed to leverage the catastrophic forgetting over the forgetting data by fine-tuning the original model only on the remaining dataset. Furthermore, the gradient ascent method involves training the original model with a reverse gradient (i.e., increasing the training loss) for the data to be scrubbed. Amnesiac unlearning approach~\cite{graves2021amnesiac} was proposed to fine-tune the model with randomly labeled forgetting samples along with unchanged remaining data. Boundary unlearning~\cite{chen2023boundary} shifted focus from the parameter space to the decision space, facilitating the rapid elimination of the target class. \citet{chundawat2023can} introduced a teacher-student framework employing distillation techniques to differentiate between beneficial and detrimental influences using good and bad teachers to enhance the learning process. SCRUB~\cite{kurmanji2024towards} utilized the positive distillation and negative distillation on the remaining data and targeted data, respectively, thereby obviating the need for bad teachers. Another line of work adopted an additive Gaussian noise to perturb the parameters based on the Fisher information matrix on targeted data~\cite{golatkar2020eternal}. Furthermore, some works leveraged the influence function approach to characterize the change in parameters if a training point is removed from the training loss~\cite{koh2017understanding}. Nonetheless, existing approximate unlearning methods still have some disadvantages including inadequate methodological foundation, high computational complexity, and the demand for substantial memory capacity during the unlearning process.

\paragraph{Siamese networks.} Siamese networks~\cite{bromley1993signature} are a group of versatile models widely employed across a diverse range of applications, including object tracking~\cite{bertinetto2016fully}, face verification~\cite{taigman2014deepface}, one-shot learning~\cite{koch2015siamese}, and self-supervised learning~\cite{chen2020simple, chen2021exploring}. These networks, characterized by weight-sharing and the comparison of entities through two inputs, have attracted significant interest due to their ability to learn meaningful representations and extract knowledge without explicit supervision. Recent advancements in self-supervised learning have particularly harnessed Siamese networks to facilitate the acquisition of valuable insights. Given these works, the Siamese network offers unique advantages in unlearning by enabling a direct comparison between representations of forgetting and retained data, facilitating targeted knowledge removal. 
\section{Method}
\label{sec:method}
\subsection{Preliminaries and Notation}
Let the complete training dataset be $\D = \{\x^{(i)}, y^{(i)} \}_{i=1}^N \subseteq \mathcal{X} \times \mathcal{Y}$, which consists of $N$ input samples $\x^{(i)} \in \mathcal{X}$ with corresponding label $y^{(i)} \in \mathcal{Y}$. As a subset of the training dataset, we denote $\mathcal{D}_f \subseteq \mathcal{D}$ as the forgetting data that consists of $N_f$ data samples and let $\mathcal{D}_r = \mathcal{D} \setminus \mathcal{D}_f$ denote the remaining dataset with the size of $N_r$. 

The DNN model is denoted as $f_{\bomega}$, where $\bomega$ represents the trainable parameters. Given an input $\x$, $f_{\bomega}(\x)$ produces the logits $\boldsymbol{l}$. We denote the original DNN model trained on $\mathcal{D}$ as $f_{\bomega_0}$ parameterized by $\bomega_0$. 
The original DNN model trained on dataset $\mathcal{D}$ is denoted as $f_{\bomega_0}$ with parameters $\bomega_0$ and the model $f_{\bomega^*}$ is trained on remaining data $\mathcal{D}_r$. We aim to unlearn the knowledge of $\mathcal{D}_f$ from $f_{\bomega_0}$ by updating the parameters $\bomega_0$ to $\tilde{\bomega}$ that $f_{\tilde{\bomega}}$ is expected to be the as same as $f_{\bomega^*}$. Note that achieving this objective does not necessarily mean that the updated parameters $\tilde{\bomega}$ will be similar to $\bomega^*$, as models with different parameter configurations can exhibit the same behavior~\cite{thudi2022necessity}.

\subsection{Knowledge Vaporization and Concentration}
Before introducing the proposed method, we revisit insights gained from the visualization of the logits of augmented data samples output by the retained model. 
As presented in Figure~\ref{fig:intro_fig} and discussed before, the augmented views of one forgetting data are much more well-dispersed in the embedding space, while the augmented views of one remaining data sample are concentrated. This observation matches the intrinsic essence of unlearning: erasing the knowledge learned from the forgetting data. 

To imitate the behavior of the retrained model, we aim to disperse encoded representations of augmented views, thereby enhancing randomness and entropy within these representations. This process, akin to physical vaporization, is termed \emph{knowledge vaporization} and is formalized by the following expression
\begin{align}
    \max\limits_{\boldsymbol{\omega}} \mathbb{E}_{\mathcal{T}}\left[\frac{1}{N_f}\sum\limits_{\x \in \mathcal{D}_f}\|f_{\boldsymbol{\omega}}(\mathcal{T}(\x)) - \eta_{\x} \|_2^2\right], \label{eq:theoretical_1}
\end{align}
where $\|\cdot \|_2$ denotes $\ell_2$-norm and $\mathcal{T}$ is the data augmentation. In \eqref{eq:theoretical_1}, we introduce $\eta_{\x}$ as the underlying representation for each data point $\x$ and all the encoded representations of augmented views $\mathcal{T}(\x)$ converge to $\eta_{\x}$. This assumption is widely adopted in supervised and self-supervised learning, which implies that a well-trained model can learn a meaningful and stable representation for each data sample. Intuitively, the optimization problem~\eqref{eq:theoretical_1} can be interpreted as moving encoded representations away from $\eta_{\x}$ to facilitate the evaporation of learned knowledge.

However, solely applying knowledge vaporization to forgetting data can impair the model utility for remaining data. On the other hand, it is crucial to recognize the overlap in knowledge learned from both forgetting and remaining data. Because of this overlap, the generalization of specific remaining data can also maintain the model performance on forgetting data. Therefore, we introduce the complementary concept of \emph{knowledge concentration}, formulated to minimize the expected distance:
\begin{align}
    \min\limits_{\boldsymbol{\omega}} \mathbb{E}_{\mathcal{T}}\left[\frac{1}{N_r}\sum\limits_{\x \in \mathcal{D}_r}\|f_{\boldsymbol{\omega}}(\mathcal{T}(\x)) - \eta_{\x} \|_2^2\right].
\end{align}

\begin{figure}[t]
\centering
\includegraphics[width=0.99\columnwidth]{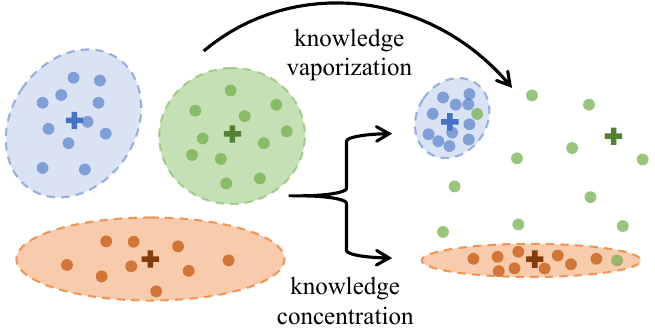} 
\caption{Illustration of the concepts of knowledge concentration and knowledge vaporization. Dots of varying colors depict logits output from augmented views of different data samples. Knowledge concentration leads to more concentrated logits (e.g., blue and orange dots), while knowledge vaporization results in dispersed logits for a single data sample (e.g., green dots). }
\label{fig:concept}
\end{figure}

Knowledge vaporization and concentration are adversarial objectives that cooperate in approximate unlearning. As illustrated in Figure~\ref{fig:concept}, knowledge vaporization disperses the logits of augmented views of forgetting data samples, while knowledge concentration maintains concentration for the remaining data. They serve the dual purpose of eliminating and preserving knowledge, working together to achieve selective knowledge deletion.
\subsection{Siamese Machine Unlearning}
\begin{figure*}[t]
\centering
\includegraphics[width=1.8\columnwidth]{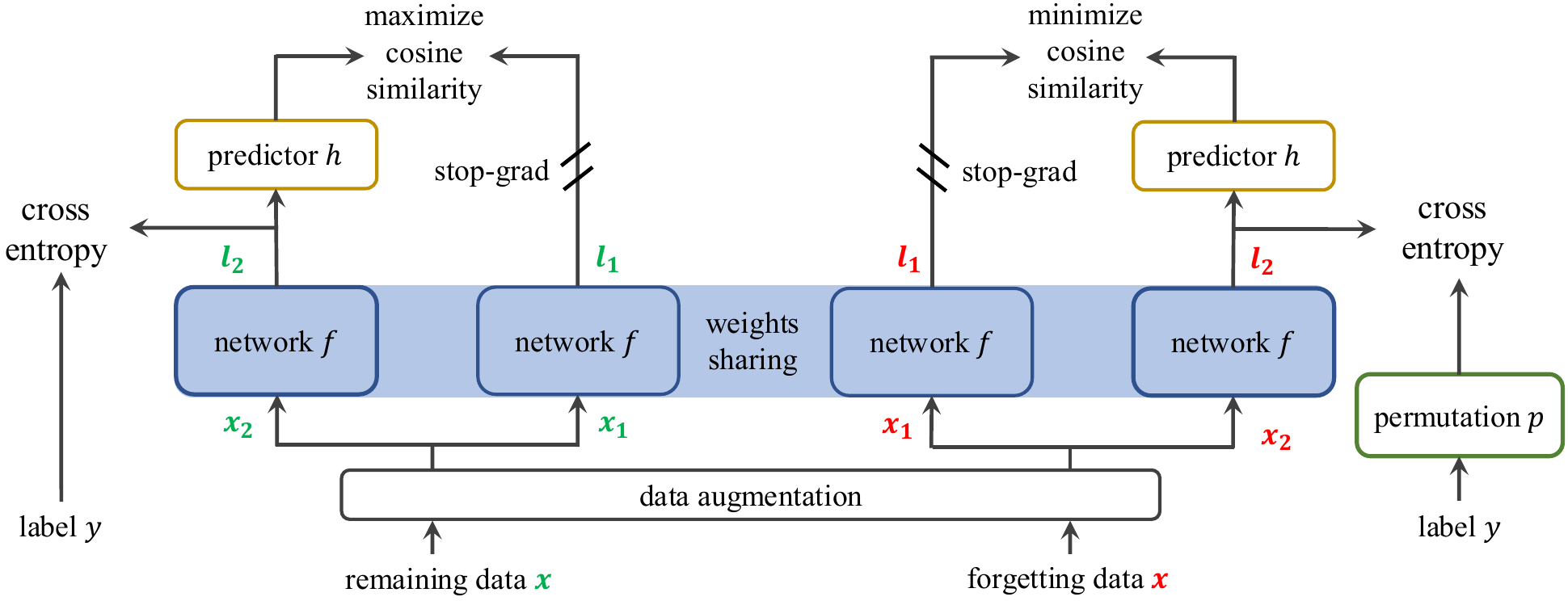} 
\caption{The proposed Siamese network for unlearning. Two augmented views of one data point are processed by a network $f$ with sharing weights. One side of the logit is input to a prediction MLP $h$, while a stop-gradient operation is applied to the other side. Knowledge concentration for remaining data: The model maximizes the similarity between both sides and optimizes the cross entropy with the label. Knowledge vaporization for forgetting data: The model minimizes the similarity between both sides and optimizes the cross entropy with the permuted labels.}
\label{fig:model}
\end{figure*}
Based on the concepts of knowledge vaporization and concentration, our objective is to develop a particular network architecture to not only achieve these unlearning goals but also cater to additional requirements grounded in practical scenarios.
\begin{itemize}
    \item \textbf{Limited access to remaining data:} In practical settings, access to the remaining dataset is restricted. Hence, unlearning should solely depend on a few remaining data samples $\mathcal{S}_r\subset \mathcal{D}_r$ rather than the entire remaining data $\mathcal{D}_r$, which has been adopted in seminal works~\cite{chundawat2023zero,10113700}
    \item \textbf{Memory efficiency:} Unlike unlearning methods based on teacher-student frameworks, only the original model will be processed without the need for additional neural networks throughout the unlearning process. 
\end{itemize}
To meet these requirements, we leverage data augmentation and design an unlearning framework based on Siamese networks. Specifically, our method, illustrated in Figure~\ref{fig:model}, takes two randomly augmented views, $\boldsymbol{x}_1$ and $\boldsymbol{x}_2$, as input from an original data sample $\boldsymbol{x}$. These views are processed by the original neural network $f$ (e.g., VGG~\cite{simonyan2014very} or ResNet~\cite{he2016deep}), with shared weights between the two views. To match the output of one view to the other view, a prediction MLP head, denoted as $h$, transforms the output. We define the distance as the negative cosine similarity between the two output logits, $\boldsymbol{p}_1 = h(f_{\boldsymbol{\omega}}(\boldsymbol{x}_1))$ and $\boldsymbol{l}_2 = f_{\boldsymbol{\omega}}(\boldsymbol{x}_2)$:
\begin{align}
    d(\boldsymbol{p}_1, \boldsymbol{l}_2) = -\frac{\boldsymbol{p}_1^T \boldsymbol{l}_2}{\|\boldsymbol{p}_1\|_2 \|\boldsymbol{l}_2\|_2 },
\end{align}
where $d(\cdot)$ is equivalent to the cosine distance between $\boldsymbol{p}_1$ and $\boldsymbol{l}_2$. To achieve knowledge vaporization, which maximizes the distance of logits between two augmentation views $\boldsymbol{x}_1$ and $\boldsymbol{x}_2$, we minimize the loss term $\mathcal{L}_{KV}(\boldsymbol{\omega}; \x)$, symmetrically defined for $\boldsymbol{x}_1$ and $\boldsymbol{x}_2$, 
\begin{align}
\label{eq:kv}
    \mathcal{L}_{KV}(\boldsymbol{\omega}; \x) & = - \frac{1}{2}\left[d(\boldsymbol{p}_1, \text{sg}(\boldsymbol{l}_2))  + d(\boldsymbol{p}_2, \text{sg}(\boldsymbol{l}_1)) \right],
\end{align}
where $\text{sg}(\cdot)$ denotes the stop-gradient technique on the output logits. 
As presented in seminal works for self-supervised learning such as Simsiam~\cite{chen2021exploring}, the stop-gradient stabilizes the optimization process and prevents the collapse of representations. Thus, we implement knowledge vaporization on the original model by minimizing $\mathcal{L}_{KV}$ for every forgetting data $\x \in \mathcal{D}_f$.

Knowledge concentration can be achieved oppositely. In particular, for every remaining data sample $\x \in \mathcal{S}_r$, we formulate the concentration loss $\mathcal{L}_{KC}(\boldsymbol{\omega}; \x)$ as:
\begin{align}
    \mathcal{L}_{KC}(\boldsymbol{\omega}; \x) & = \frac{1}{2}\left[d(\boldsymbol{p}_1, \text{sg}(\boldsymbol{l}_2))  + d(\boldsymbol{p}_2, \text{sg}(\boldsymbol{l}_1)) \right].
\end{align}

However, only optimizing $\mathcal{L}_{KV}(\bomega; \x)$ and $\mathcal{L}_{KC}(\bomega; \x)$ may lead to outputs of the Siamese network that are not conducive for prediction tasks. To align the output for prediction and guide the optimization of knowledge vaporization and concentration, we concurrently optimize the Siamese network with the task loss. As presented in Figure~\ref{fig:model}, we introduce the cross-entropy term for the logits without stop-gradient on two augmented views $\x_1$ and $\x_2$. For a remaining data $\x \in \mathcal{S}_r$, this term in knowledge concentration is denoted as $\text{SCE}(\x, y)$ and symmetrically defined for $\x_1$ and $\x_2$,
\begin{align}
    \text{SCE}(\x, y) = \frac{1}{2}[\text{CE}(\boldsymbol{l}_1, y) + \text{CE}(\boldsymbol{l}_2, y)].
\end{align}
Nevertheless, for the forgetting data, directly using the true label $y$ or random labels may not be suitable. To this end, we develop an adaptive label permutation for knowledge vaporization, which will be introduced in the next subsection.

\paragraph{Adaptive label permutation.} Unlearning impacts the model performance for each class based on the proportion of unlearned data within each class. In extreme cases, unlearning the entire data for a specific class would result in the model failing on this class. Conversely, unlearning a single data point within a class would have minimal impact on the model's utility for that class.

To achieve that, we introduce the cross-entropy term on $\mathcal{D}_f$ with an adaptive label permutation $p$. For the $k$-th class, the permutated label $p(y)$ should depend on the ratio of unlearned data samples in each class of the entire dataset. To achieve this goal, we adopt a common technique for local differential privacy, random response~\cite{warner1965randomized}, to permutate the label according to the ratio of unlearned data within each class. Let $\boldsymbol{r} = (r_1, r_2, \dots, r_K)$ denote the vector of ratios of unlearned data within each class, where $0\leq r_k\leq 1, k\in {1, 2, \dots, K}$. Given $\boldsymbol{r}$, for $k \in {1, 2, \dots, K}$, the transmit probability is given by
\begin{align}\label{eq: RR}
    \Pr[p(k) = v] = \left\{  
\begin{array}{lr}
    \frac{r_k^{-1}}{r_k^{-1}+k-1}, & \text{if } v = k,    \\
    \frac{1}{r_k^{-1}+k-1} , &\text{if } v \not = k. 
\end{array}
\right. 
\end{align} 
When $r_k = 1$, the permuted label is distributed randomly across all classes. Conversely, when $r_k = 0$, no permutation occurs, aligning with our initial expectations.

Having established the objective terms for knowledge vaporization and concentration, the ultimate training objective of Siamese unlearning is as follows:
\begin{align}
    \min\limits_{\boldsymbol{\omega}} & \frac{1}{N'_r} \sum\limits_{\x \in \mathcal{S}_r}\left[ \mathcal{L}_{KC} \left(\boldsymbol{\omega}, \x\right) + \lambda\text{SCE}\left(\x, y\right) \right]\\
    & + \frac{1}{N_f} \sum\limits_{\x \in \mathcal{D}_f}\left[ \mathcal{L}_{KV} \left(\boldsymbol{\omega}, \x\right) + \lambda\text{SCE}\left(\x, p(y)\right) \right],
\end{align}
where $\lambda \geq 0$ is a hyperparameter and $N'_r$ denotes the size of few remaining data subset $\mathcal{S}_r$.
\section{Experiments}
\label{sec:4_experiments}
\subsection{Experimental Settings}
\paragraph{Datasets and model architectures.} In the experiments, we adhere to seminal works and utilize CIFAR-10 and CIFAR-100 datasets~\cite{krizhevsky2009learning} as our benchmark datasets. We consider three architectures: VGG16-BN~\cite{simonyan2014very}, ResNet18, and ResNet50~\cite{he2016deep}. Moreover, we adopt three different data augmentations: \romannumeral1) \emph{simple} augmentation consisting of crop and random horizontal flip, \romannumeral2) \emph{contrastive}~\cite{chen2020simple} data augmentation that is commonly used in contrastive learning, and \romannumeral3) \emph{Cutout}~\cite{devries2017improved} data augmentation\footnote{More implement details and experimental results are provided in Appendices and the source code of the proposed Siamese unlearning is available at: https://github.com/SongjieXie/SiamUnlearning}. 
\begin{figure*}
\centering
\begin{subfigure}{0.69\columnwidth}
  \includegraphics[width=0.97\columnwidth]{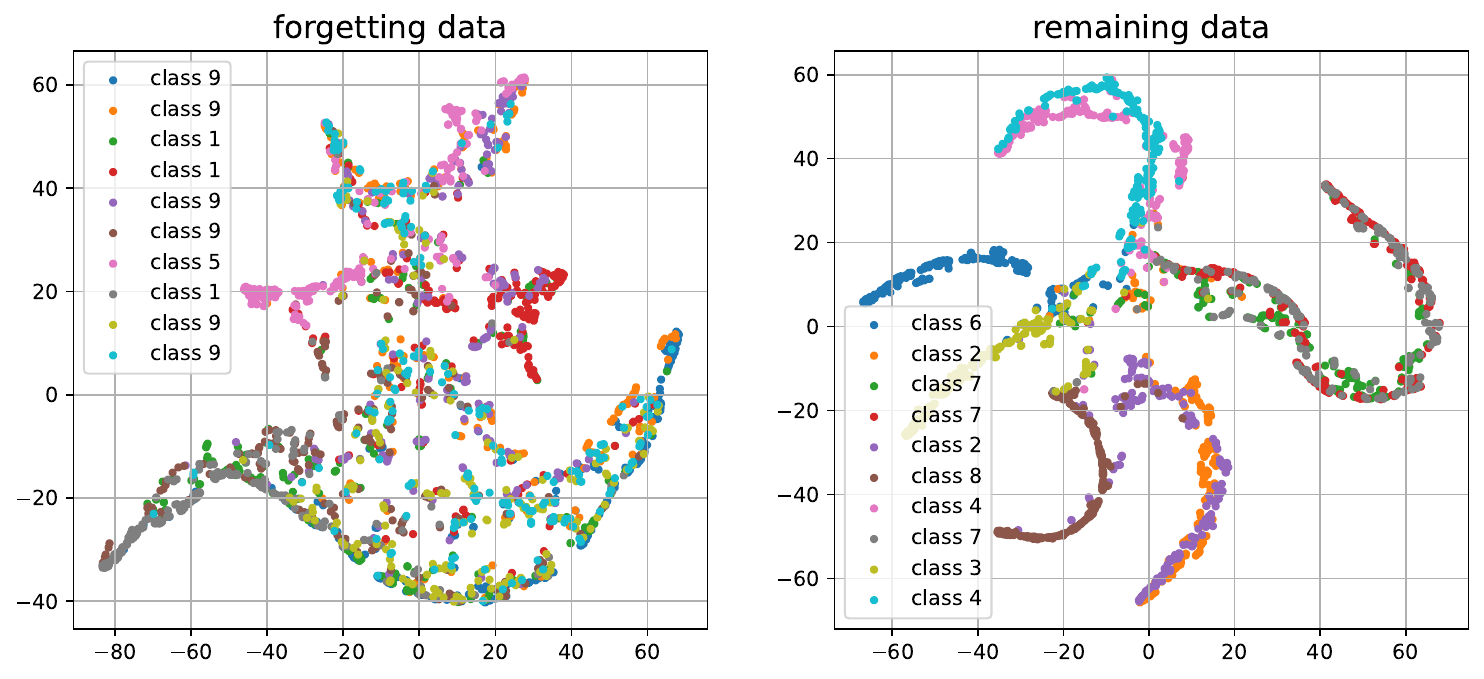}
  \caption{Full-class unlearning}
\end{subfigure}%
\begin{subfigure}{0.69\columnwidth}
  \includegraphics[width=0.97\columnwidth]{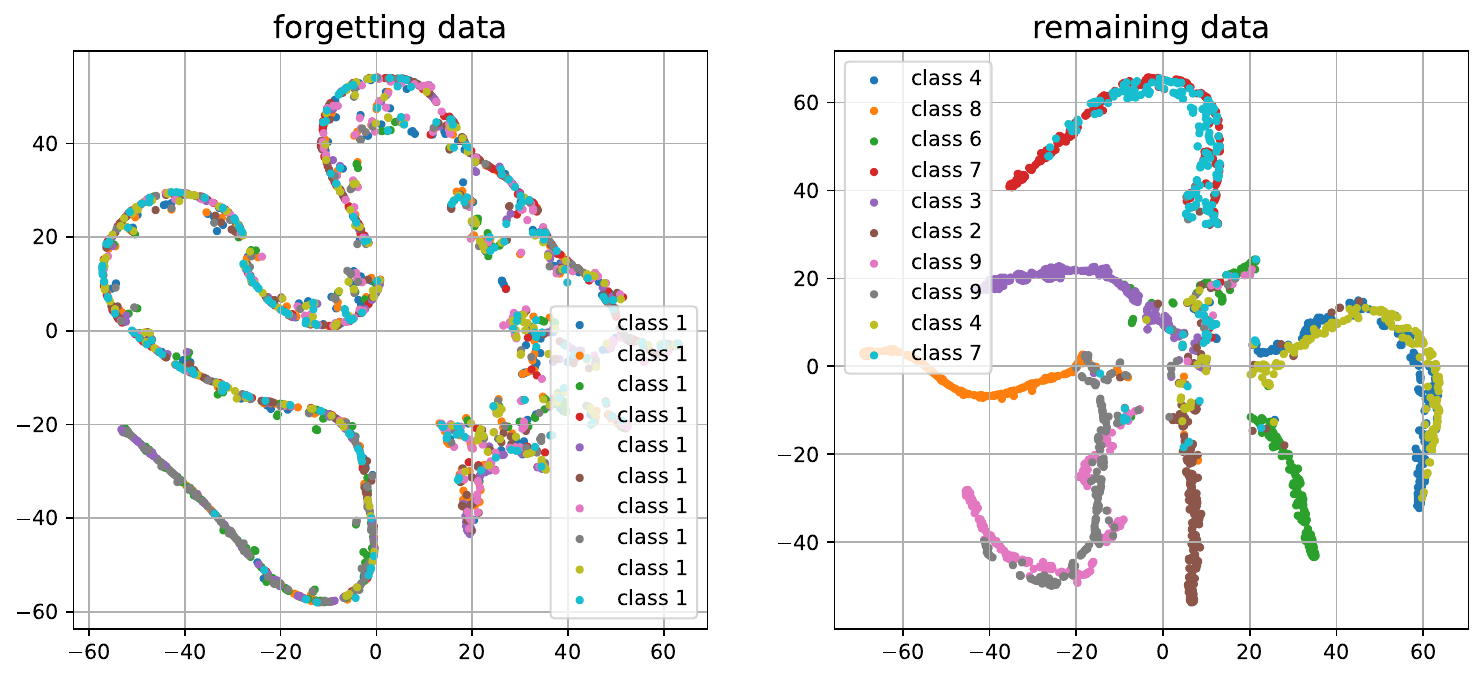}
  \caption{Sub-class unlearning}
\end{subfigure}
\begin{subfigure}{0.69\columnwidth}
  \includegraphics[width=0.97\columnwidth]{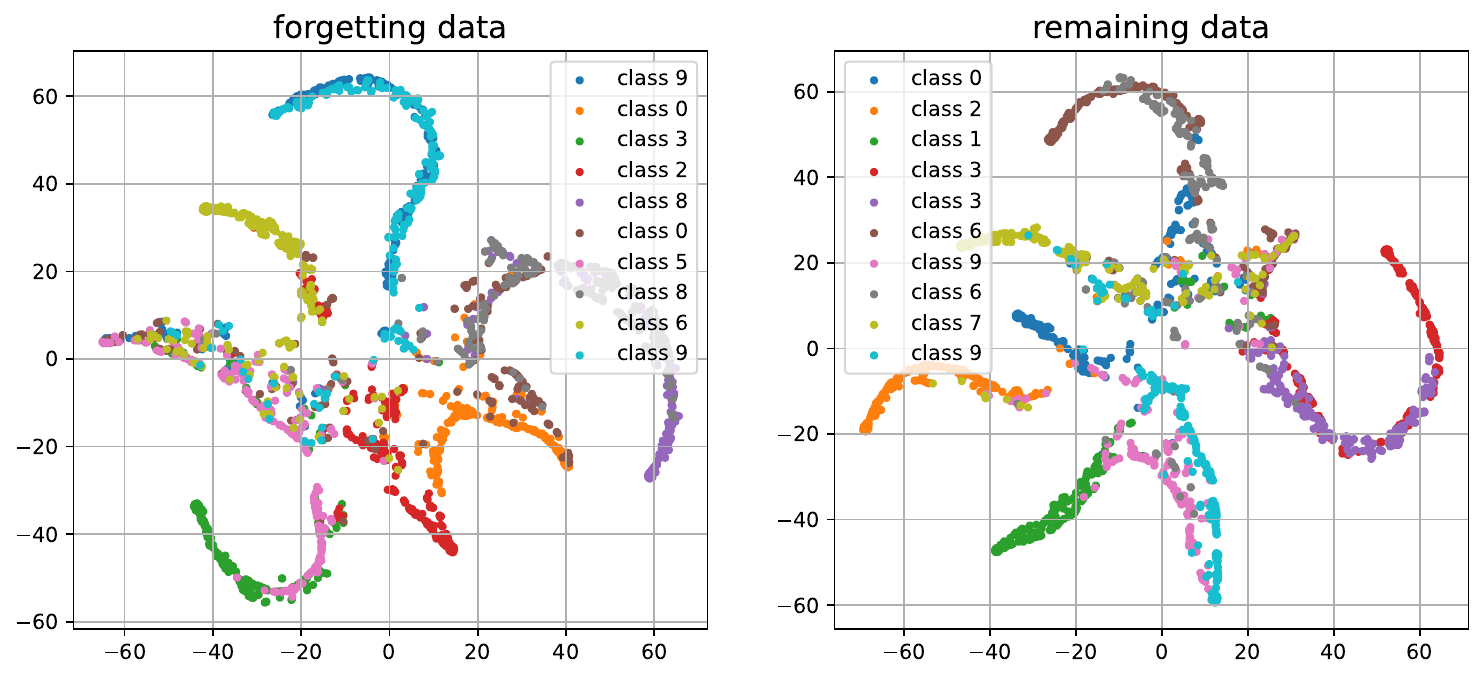}
  \caption{Random unlearning}
\end{subfigure}
\caption{Visualization of logits from Siamese unlearned models for scenarios: (a) full-class, (b) sub-class, and (c) random forgetting.}
\end{figure*}
\paragraph{Baseline methods.} We benchmark state-of-the-art unlearning baseline methods for comparison: 
\romannumeral1) \emph{Original} is the original model $\boldsymbol{\omega}_0$ without unlearning. \romannumeral2) \emph{Retrain} denotes the model trained on only the remaining data $\mathcal{D}_r$. \romannumeral3) \emph{Finetune}~\cite{warnecke2021machine} fine-tunes $\boldsymbol{\omega}_0$ only with remaining dataset $\mathcal{D}_r$, utilizing catastrophic forgetting to achieve unlearning. \romannumeral4) \emph{NegGrad}~\cite{golatkar2020eternal} fine-tunes $\boldsymbol{\omega}_0$ with reverse gradient steps on the forgetting dataset $\mathcal{D}_f$. \romannumeral5) \emph{RandLab} also apply finetuning on $\boldsymbol{\omega}_0$ with randomly labeled $\mathcal{D}_f$. There are two baselines based on teacher-student frameworks, \romannumeral6) knowledge distillation from a bad teacher (\emph{BadT})~\cite{chundawat2023can} and \romannumeral7) scalable remembering and unlearning unbound (\emph{SCRUB})~\cite{kurmanji2024towards}. \romannumeral8) Selective Synaptic Dampening (\emph{SSD})~\cite{foster2024fast} selectively modifies $\boldsymbol{\omega}_0$ with high influence from $\mathcal{D}_f$. \romannumeral9) \emph{Amnesiac}~\cite{graves2021amnesiac} fine-tunes $\boldsymbol{\omega}_0$ using randomly labeled $\mathcal{D}_f$ along with $\mathcal{D}_r$. 
\paragraph{Unlearning scenarios and metrics.} We evaluate the unlearning methods across three typical unlearning scenarios: \romannumeral1) \emph{Full-class forgetting}, where the full classes from the dataset shall be unlearned, \romannumeral2) \emph{Sub-class forgetting}, where a related subset from a class is learned, and \romannumeral3) \emph{Random forgetting}, where a subset is sampled uniformly from the entire training dataset $\mathcal{D}$. In all three scenarios, the performance of evaluated methods is measured by the following metrics: \romannumeral1) accuracy on $\mathcal{D}_r$ (\emph{$\text{Acc}_{\mathcal{D}_r}$}), \romannumeral2) accuracy on $\mathcal{D}_f$ (\emph{$\text{Acc}_{\mathcal{D}_f}$}), \romannumeral3) accuracy on test dataset (\emph{TA}), and membership inference attack ratio (\emph{MIA})~\cite{song2021systematic}. In full / sub-class unlearning, 
we denote the test accuracy on samples of the forgetting classes and the remaining classes as \emph{$\text{TA}_{\mathcal{D}_r}$} and \emph{$\text{TA}_{\mathcal{D}_f}$}, respectively.
\subsection{Experimental Results}
\begin{figure}
\centering
\includegraphics[width=0.95\columnwidth]{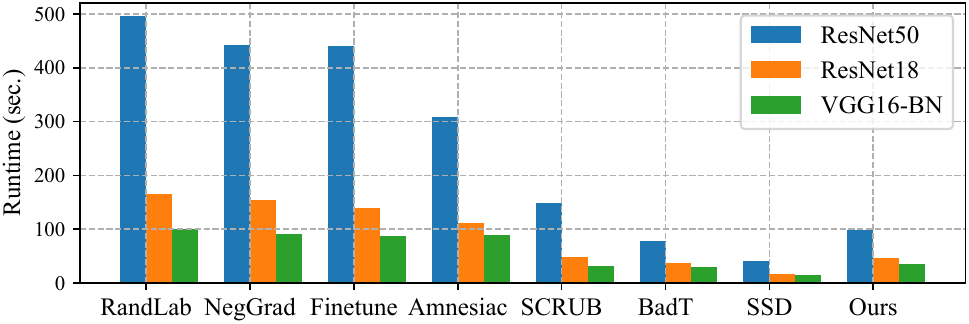} 
\caption{Runtime of the evaluated methods on CIFAR-100 dataset with different backbones.}
\label{fig:runtime}
\end{figure}

To validate the efficacy of the proposed method, we consider three unlearning scenarios and apply the proposed Siamese unlearning approach to the CIFAR-10 dataset with contrastive augmentation. Subsequently, we visualized the output logits corresponding to forgotten and retained data, as depicted in Figure~\ref{fig:exp_tsne}. The visualizations closely match the insights gained from similar visualizations of retrained models, as illustrated in Figure~\ref{fig:intro_fig}. Notably, the augmented views of forgotten data points exhibit dispersion across the embedding space, whereas the augmented retained data points maintain dense and coherent clusters associated with their respective classes. This observation underscores the ability of the Siamese unlearning method to align the output logits of unlearned models more closely with those of retrained models, thereby selectively eliminating learned knowledge from forgotten data.
\paragraph{Full-class unlearning.} Table~\ref{tab:full-cifar10} and Table~\ref{tab:full-cifar100} show the performance of different unlearning methods in the full-class scenario, utilizing VGG16-BN and ResNet50 architectures on CIFAR-10 and CIFAR-100 datasets. Superior performance is indicated by higher accuracy metrics on $\mathcal{D}_r$ (i.e., $\text{Acc}(\mathcal{D}_r)$ and $\text{TA}(\mathcal{D}_r)$), or lower accuracy on $\mathcal{D}_f$ (i.e., $\text{Acc}(\mathcal{D}_f)$ and $\text{TA}(\mathcal{D}_f)$). Our method achieves the best performance on the training and test data of $\mathcal{D}_r$ and the lowest accuracy on $\mathcal{D}_f$, which means it effectively erases the knowledge of $\mathcal{D}_f$ and maintains the utility on $\mathcal{D}_f$. Notably, our method achieves a perfect $0.00$ MIA score on both datasets, indicating that no data points from forgotten classes are recognized as training data, thus providing empirical evidence of forgetting.
\begin{table}[t]
	\centering
	\caption{Full-class unlearning performance on CIFAR-10 dataset using VGG16-BN with the simple augmentation.  The unlearned classes are \texttt{automobile}, \texttt{cat}, and \texttt{truck}. The best and second-best results are bolded and underlined, respectively}
	\label{tab:full-cifar10}
	\resizebox{0.95\linewidth}{!}{
				\begin{tabular}{lccccc}
					\toprule
					\textbf{Approach} &$\text{Acc}_{\mathcal{D}_r} \uparrow$ & $\text{Acc}_{\mathcal{D}_f}\downarrow$ & $\text{TA}_{\mathcal{D}_r}\uparrow$& $\text{TA}_{\mathcal{D}_f}\downarrow$& $\text{MIA}\downarrow$\\
					\midrule
					Original     & $99.99$          &$99.99$           &$93.63$            & $94.91$ & $95.25$\\ 
					Retrain      & $100.00$         &$0.00$            &$93.79$            & $0.00$  & $19.64$\\ 
                    \midrule
                    Finetune     & $99.28$ & $\mathbf{0.00}$ &$93.26$ & $\mathbf{0.00}$ & $6.05$\\
                    NegGrad &  $98.62$ & $0.08$ & $91.49$ & $0.16$ & $14.85$\\
                    Amnesiac & $98.55$ &  $\mathbf{0.00}$ &$\underline{93.70}$ & $\mathbf{0.00}$ & $22.66$\\
                    RandLab & $93.66$ & $0.11$ & $80.27$ & $\underline{0.03}$ & $6.57$\\
                    BadT & $\underline{99.47}$ & $2.97$ & $92.83$ & $11.30$ & $\underline{0.04}$\\
                    SCRUB & $99.17$ & $\mathbf{0.00}$ & $93.58$ & $\mathbf{0.00}$ & $20.98$\\
                    SSD &$99.41$ & $3.25$ & $93.10$ & $\mathbf{0.00}$ & $7.26$\\
                    
                    \textbf{Ours} & $\mathbf{99.98}$ & $\underline{0.01}$ & $\mathbf{94.29}$ & $\mathbf{0.00}$ &  $\mathbf{0.00}$\\
					\bottomrule
				\end{tabular}	
	}
\end{table}
\begin{table}[t]
	\centering
	\caption{Full-class unlearning performance on CIFAR-100 dataset using ResNet50 with the contrastive augmentation.  The unlearned classes are \texttt{dolphin}, \texttt{seal}, and \texttt{trout}.}
	\label{tab:full-cifar100}
	\resizebox{0.95\linewidth}{!}{
				\begin{tabular}{lccccc}
					\toprule
					\textbf{Approach} &$\text{Acc}_{\mathcal{D}_r} \uparrow$ & $\text{Acc}_{\mathcal{D}_f}\downarrow$ & $\text{TA}_{\mathcal{D}_r}\uparrow$& $\text{TA}_{\mathcal{D}_f}\downarrow$& $\text{MIA}\downarrow$\\
					\midrule
					Original     & $99.56$ & $99.74$ & $75.94$ & $83.53$ & $94.13$\\ 
					Retrain      & $96.37$ & $0.00$   & $68.16$ & $0.00$ & $20.40$\\ 
                    \midrule
                    Finetune & $91.90$ & $5.86$ & $70.70$ & $4.38$ & $15.53$ \\
                    NegGrad & $90.39$ & $0.22$ & $69.61$ & $\mathbf{0.00}$  & $16.13$\\
                    RandLab & $86.75$ & $\underline{0.07}$ & $65.89$ & $\mathbf{0.00}$ & $16.80$\\
                    BadT &  $\underline{99.33}$ & $19.90$ & $\mathbf{74.80}$ & $11.06$ & $\mathbf{0.00}$\\
                    SCRUB & $98.82$ & $68.48$ & $\underline{74.21}$& $42.28$ & $24.87$\\
                    SSD & $98.37$ & $\mathbf{0.00}$ & $73.37$& $\mathbf{0.00}$ & $3.40$ \\
                    Amnesiac& $92.04$ & $0.20$ & $68.78$ & $\mathbf{0.00}$ & $\underline{1.33}$\\
                    \textbf{Ours} & $\mathbf{99.93}$ & $\mathbf{0.00}$ & $73.96$& $\mathbf{0.00}$ & $\mathbf{0.00}$\\
					\bottomrule
				\end{tabular}	
	}
\end{table}
\paragraph{Sub-class unlearning.} 
In scenarios of sub-class unlearning, the objective remains high accuracy on $\mathcal{D}_r$, while preserving partial model utility and performance on forgetting classes through generalization of the remaining $10\%$ data. A narrower performance gap with the retrained model signifies superior performance rather than lower accuracy. Table~\ref{tab:sub-cifar10} presents contrasting cases where SSD achieves high accuracy on $\mathcal{D}_r$ but also on forgetting data, while Amnesiac and RandLab achieve zero accuracy on forgetting data but limited utility on remaining data. In comparison, our method achieves relatively high performance on $\mathcal{D}_r$ with a minimal performance gap in $\text{Acc}(\mathcal{D}_f)$ and $\text{TA}(\mathcal{D}_f)$ when compared to the retrained model. Notably, the MIA scores of our method are even lower than those of retrained models, primarily due to knowledge concentration and vaporization rendering the output of unlearned models indistinguishable and resilient against MIA attacks.
\begin{table}[t]
	\centering
        \caption{Sub-class unlearning performance on CIFAR-10 dataset using VGG16-BN with the simple augmentation. The forgetting data is 90\% training data of class \texttt{automobile}. The results of $\text{Acc}_{\mathcal{D}_f}$ and $\text{TA}_{\mathcal{D}_f}$ are given by $a_{(b)}$, where $b$ denotes the performance gap with the retrain model.}
	\label{tab:sub-cifar10}
	\resizebox{0.99\linewidth}{!}{
				\begin{tabular}{lccccc}
					\toprule
					\textbf{Approach} &$\text{Acc}_{\mathcal{D}_r} \uparrow$ & $\text{Acc}_{\mathcal{D}_f}$ & $\text{TA}_{\mathcal{D}_r}\uparrow$& $\text{TA}_{\mathcal{D}_f}$& $\text{MIA}\downarrow$\\
					\midrule
					Original     & $99.99$ &$100.00$ &$93.53$ & $97.69$ & $97.33$\\ 
					Retrain      & $99.99$ & $79.82$ & $92.38$ & $80.98$ & $61.49$\\ 
                    \midrule
                    Finetune & $99.59$ & $97.97_{(-18.15)}$ & $92.37$ & $94.16_{(13.18)}$ & $91.16$\\
                    NegGrad & $90.08$ & $\underline{80.65_{(0.83)}}$ & $86.03$ & $\mathbf{78.84}_{(-2.14)}$ & $\underline{43.87}$\\
                    RandLab & $90.04$ & $0.00_{(-79.82)}$ & $86.19$ & $0.00_{(-80.98)}$ & $\mathbf{0.00}$\\
                    BadT & $99.23$ & $99.35_{(19.53)}$ & $92.30$ & $96.03_{(15.05)}$ & $\mathbf{0.00}$\\
                    SCRUB & $\underline{99.93}$ & $96.79_{(16.97)}$ & $92.15$ & $91.56_{(10.58)}$ & $\mathbf{0.00}$\\
                    SSD &$\mathbf{99.99}$ & $100.00_{(20.18)}$ & $\mathbf{93.49}$ & $97.47_{(16.49)}$ & $91.18$ \\
                    Amnesiac&  $98.12$ & $0.00_{(-79.82)}$ & $92.00$ & $0.00_{(-80.98)}$ & $\mathbf{0.00}$\\
                    \textbf{Ours} & $99.43$ & $\mathbf{79.54_{(-0.28)}}$ & $\underline{92.49}$ & $\underline{75.30_{(-5.68)}}$ & $\mathbf{0.00}$\\
					\bottomrule
				\end{tabular}	
	}
\end{table}
\begin{table}[t]
	\centering
        \caption{Sub-class unlearning performance on CIFAR-100 dataset using ResNet18 with the contrastive augmentation. The forgetting data is 90\% training data of class \texttt{dolphin}.}
	\label{tab:sub-cifar100}
	\resizebox{0.99\linewidth}{!}{
				\begin{tabular}{lccccc}
					\toprule
					\textbf{Approach} &$\text{Acc}_{\mathcal{D}_r} \uparrow$ & $\text{Acc}_{\mathcal{D}_f}$ & $\text{TA}_{\mathcal{D}_r}\uparrow$& $\text{TA}_{\mathcal{D}_f}$& $\text{MIA}\downarrow$\\
					\midrule
					Original     & $99.33$ & $99.61$ & $74.91$ & $89.49$ & $96.44$\\ 
					Retrain      & $91.23$ & $27.88$ & $71.86$ & $24.99$ & $12.44$\\ 
                    \midrule
                    Finetune & $95.55$ & $82.36_{(54.48)}$ & $72.01$ & $72.00_{(47.01)}$ & $60.44$\\
                    NegGrad & $64.60$ & $\mathbf{19.71_{(-8.17)}}$ & $58.34$ & $9.02_{(-15.97)}$ & $12.44$\\
                    RandLab & $61.65$ & $3.87_{(-24.01)}$ & $53.02$ & $0.00_{(-24.99)}$ & $13.78$\\
                    BadT & $96.58$ & $46.70_{(18.82)}$ & $\mathbf{73.94}$ & $\mathbf{34.00_{(9.01)}}$ & $10.44$\\
                    SCRUB & $\mathbf{98.65}$ & $94.22_{(66.34)}$ & $\underline{73.26}$ & $80.00_{(55.01)}$ & $70.22$\\
                    SSD & $72.57$ & $0.00_{(-27.88)}$ & $54.09$& $0.00_{(-24.99)}$ & $9.33$ \\
                    Amnesiac& $96.61$ & $48.88_{(21.00)}$ & $72.70$ & $49.00_{(24.01)}$ & $\mathbf{2.22}$\\
                    \textbf{Ours} & $\underline{96.93}$ & $\underline{45.57_{(17.69)}}$ & $72.38$ & $\underline{31.00_{(10.01)}}$ & $\underline{5.11}$\\
					\bottomrule
				\end{tabular}	
	}
\end{table}
\paragraph{Random unlearning.} 
Tables~\ref{tab:random-cifar10} and \ref{tab:random-cifar100} present the performance of the compared methods in random unlearning scenarios. In contrast to full-class and sub-class scenarios, we only consider the test accuracy on the entire test dataset $\mathcal{D}_f$, reflecting the preserved utility after unlearning $10\%$ of training data samples. As the performance on $\mathcal{D}_f$ also depends on the generalization capacity of $\mathcal{D}_r$, we anticipate unlearned models to exhibit a minor accuracy gap compared to retrained models. Our method consistently outperforms the baseline methods in this context. While the MIA scores of our method are slightly higher than some baselines, they remain close to and lower than the MIA scores of retrained models.
\begin{table}[t]
	\centering
	\caption{Random unlearning performance on CIFAR-10 for 10\% training samples using VGG16-BN with the simple augmentation.}
	\label{tab:random-cifar10}
	\resizebox{0.89\linewidth}{!}{
				\begin{tabular}{lcccc}
					\toprule
					\textbf{Approach} &$\text{Acc}_{\mathcal{D}_r} \uparrow$ & $\text{Acc}_{\mathcal{D}_f}$ & $\text{TA}_{\mathcal{D}}\uparrow$& $\text{MIA}\downarrow$\\
					\midrule
					Original     &$99.99$ &$100.00$ & $94.00$ &$97.34$ \\ 
					Retrain      &$99.10$ &$99.14$ & $92.64$ &  $94.64$\\ 
                    \midrule
                    Finetune& $99.63$ & $98.18_{(-0.96)}$ & $92.41$& $89.08$\\
                    NegGrad & $98.97$ & $98.07_{(-1.07)}$ & $91.62$ & $90.60$\\
                    RandLab & $93.22$ & $77.17_{(-21.97)}$ & $86.56$ & $\underline{58.66}$\\
                    BadT &  $99.67$ & $97.21_{(-1.93)}$ & $92.24$ & $\mathbf{54.96}$\\
                    SCRUB & $\mathbf{99.99}$ & $\underline{100.00_{(0.86)}}$ & $\underline{93.67}$ & $94.38$\\
                    SSD & $99.93$ & $\mathbf{99.94_{(0.80)}}$ & $93.41$ & $96.18$\\ 
                    Amnesiac&  $96.71$ & $79.77_{(-19.37)}$ & $88.35$ & $65.08$\\
                    \textbf{Ours} & $\underline{99.98}$ & $\underline{100.00_{(0.86)}}$ & $\mathbf{93.97}$ & $62.10$\\
					\bottomrule
				\end{tabular}	
	}
\end{table}
\begin{table}[t]
	\centering
	\caption{Random unlearning performance on CIFAR-100 for 10\% training samples using ResNet50 with the Cutout augmentation.}
	\label{tab:random-cifar100}
	\resizebox{0.89\linewidth}{!}{
				\begin{tabular}{lcccc}
					\toprule
					\textbf{Approach} &$\text{Acc}_{\mathcal{D}_r} \uparrow$ & $\text{Acc}_{\mathcal{D}_f}$ & $\text{TA}_{\mathcal{D}}\uparrow$& $\text{MIA}\downarrow$\\
					\midrule
					Original     & $99.96$ & $100.00$ & $80.80$ & $95.92$ \\ 
					Retrain      & $98.25$ & $98.54$ &  $64.53$ & $89.82$\\ 
                    \midrule
                    Finetune & $97.51$ & $93.20_{(-5.34)}$ & $74.76$ & $76.34$\\
                    NegGrad & $97.92$ & $95.25_{(-3.29)}$ & $75.44$ & $88.94$\\
                    RandLab & $97.54$ & $83.59_{(-14.95)}$ & $72.23$ & $45.16$\\
                    BadT &  $97.98$ & $90.23_{(-8.31)}$ & $75.45$ & $\mathbf{42.14}$\\
                    SCRUB & ${99.55}$ & $\underline{99.94_{(1.40)}}$ & $76.63$ & $93.32$\\
                    SSD & $\mathbf{99.97}$ & $100.00_{(1.46)}$ & $\mathbf{80.77}$ & $95.20$ \\
                    Amnesiac& $97.11$ & $82.60_{(-15.94)}$ & $72.00$ & $\underline{43.68}$\\
                    \textbf{Ours} & $\underline{99.59}$ & $\mathbf{98.90_{(0.36)}}$ & $\underline{77.02}$ & $83.98$\\
					\bottomrule
				\end{tabular}	
	}
\end{table}
\paragraph{Efficiency Analysis.} Figure~\ref{fig:runtime} presents the runtimes of the evaluated methods. The proposed approach attains a relatively low runtime mainly due to the knowledge vaporization process only needs a portion of the remaining data and converges rapidly. Although BadT and SSD exhibit even shorter runtimes, our method achieves better performance in forgetting and preserving model utility. Additionally, Siamese unlearning is more memory-efficient as no additional teacher models are needed.
\subsection{Ablation Studies}
\begin{figure}
\centering
\begin{subfigure}{0.9\columnwidth}
  \includegraphics[width=0.88\columnwidth]{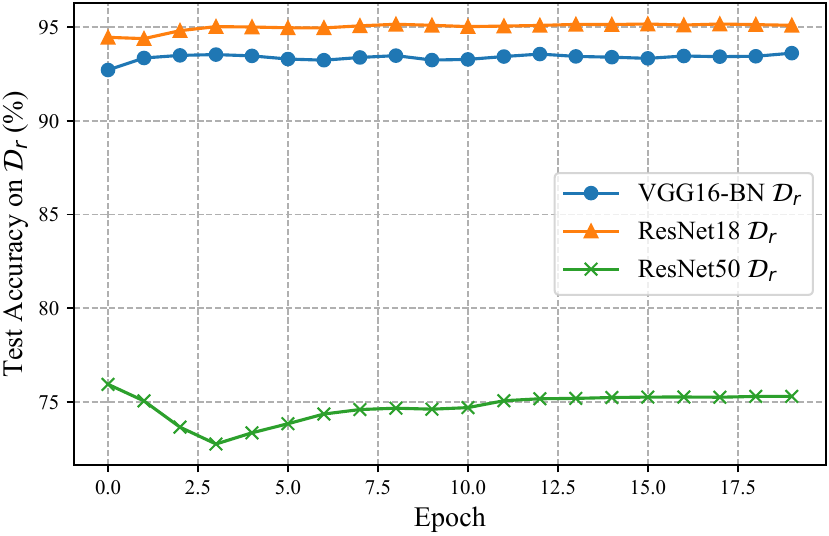}
  \caption{Knowledge concentration}
  \label{fig:ablation-kc}
\end{subfigure}%

\begin{subfigure}{0.9\columnwidth}
  \includegraphics[width=0.94\columnwidth]{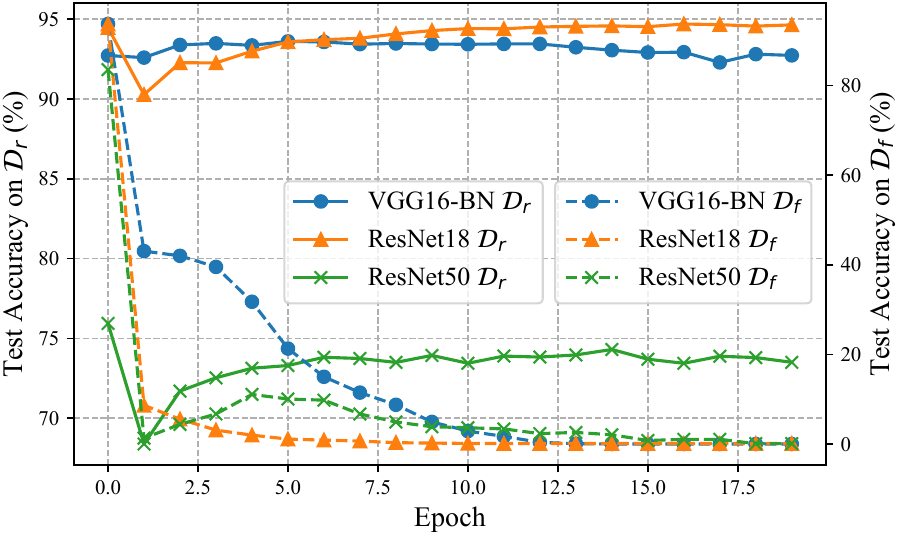}
  \caption{Knowledge concentration and vaporization}
  \label{fig:ablation-kckv}
\end{subfigure}
\caption{Test accuracy on $\mathcal{D}_f$ and $\mathcal{D}_r$ using Siamese unlearning with (a) solely knowledge concentration and (b) jointly knowledge concentration and vaporization.}
\label{fig:exp_tsne}
\end{figure}
\paragraph{Effects of knowledge concentration.} 
While the proposed knowledge vaporization aims to erase learned knowledge, the significance and necessity of knowledge concentration require validation. We conduct experiments exclusively focusing on knowledge concentration within $\mathcal{D}_r$ of CIFAR-10 using VGG16-BN and ResNet18, as well as within CIFAR-100 using ResNet50. The results, depicted in Figure~\ref{fig:ablation-kc}, show that initial knowledge concentration may temporarily impair the utility of the original models. However, soon, the updated models resume even surpassing their original performance levels (e.g., VGG16-BN and ResNet18). The concurrent implementation of knowledge vaporization and concentration is illustrated in Figure~\ref{fig:ablation-kckv}. With an increase in epochs, the values of $\text{TA}_{\mathcal{D}_f}$ consistently decrease until reaching $0.00$, while $\text{TA}_{\mathcal{D}_r}$ ascend to match the original performance after an initial decline.
\paragraph{Effects of cross-entropy terms and adaptive label permutation.} 
Table~\ref{tab:ablation-alp} displays the ablation study of the proposed knowledge vaporization and concentration (KVC), introduced cross-entropy terms (CE), and the proposed adaptive label permutation. The results indicate that without cross-entropy terms, optimizing knowledge vaporization and concentration alone leads to arbitrary representations. Conversely, optimizing only the Siamese network with cross-entropy terms, excluding knowledge vaporization and concentration, is akin to the RandLab method and yields an accuracy of $0$ on $\text{Acc}_{\mathcal{D}_f}$. Additionally, the adaptive label permutation serves to mitigate excessive damage to preserved utility resulting from generalization on the remaining data. More ablation study results on other datasets and unlearning scenarios are presented in the Appendix.

\begin{table}
	\centering
    \caption{Ablation study on CIFAR-10 under sub-class unlearning scenario. KVC denotes the loss terms of knowledge vaporization and concentration, SCE denotes the symmetric cross-entropy term, and ALP denote the adaptive label permutation.}
	\label{tab:ablation-alp}
	\resizebox{0.99\linewidth}{!}{
				\begin{tabular}{lllccccc}
					\toprule
					KVC & CE &ALP &$\text{Acc}_{\mathcal{D}_r} \uparrow$ & $\text{Acc}_{\mathcal{D}_f}$ & $\text{TA}_{\mathcal{D}_r}\uparrow$& $\text{TA}_{\mathcal{D}_f}$& $\text{MIA}\downarrow$\\
					\midrule
                    & \CheckmarkBold &  & $97.02$ & $0.00_{(-79.82)}$ & $90.63$ & $0.00_{(-80.98)}$ & $\mathbf{0.00}$\\
                    & \CheckmarkBold & \CheckmarkBold &$98.62$ &$0.00_{(-79.82)}$ & $\underline{92.23}$&  $0.00_{(-80.98)}$ & $\mathbf{0.00}$ \\
                    \CheckmarkBold &  &  & $98.73$ & $100.00_{(20.18)}$ & $90.29$ & $99.49_{(18.51)}$ & $56.56$ \\
                    \CheckmarkBold & \CheckmarkBold &  & $\underline{98.94}$ & $\underline{75.90_{(-3.92)}}$ & $91.23$ & $\underline{73.09_{(-7.89)}}$ & $\underline{00.04}$ \\
                    \CheckmarkBold & \CheckmarkBold & \CheckmarkBold & $\mathbf{99.43}$ & $\mathbf{79.54_{(-0.28)}}$ & $\mathbf{92.49}$ & $\mathbf{75.30_{(-5.68)}}$ & $\mathbf{0.00}$\\
					\bottomrule
				\end{tabular}	
	}
\end{table}

\paragraph{Effects of data augmentation.} 
In contrast to the data augmentation in contrastive learning, the proposed Siamese unlearning method is resilient and relatively unaffected by the choice of augmentations. The unlearning performance with different augmentations is shown in Table~\ref{tab:ablation-aug}. While original models trained using diverse augmentation methods display varying performances, we still can observe the consistent performance of unlearned models across various augmentation strategies. It demonstrates the resilience of the proposed methods to augmentation, ranging from basic to advanced augmentation techniques.
\begin{table}
	\centering
    \caption{The performance of original and full-class unlearned performance on CIFAR-10 classes: \texttt{automobile}, \texttt{cat}, and \texttt{truck} using ResNet18 with different data augmentation.}
	\label{tab:ablation-aug}
	\resizebox{0.9\linewidth}{!}{
				\begin{tabular}{lccccc}
					\toprule
					Aug &$\text{Acc}_{\mathcal{D}_r} \uparrow$ & $\text{Acc}_{\mathcal{D}_f}\downarrow$ & $\text{TA}_{\mathcal{D}_r}\uparrow$& $\text{TA}_{\mathcal{D}_f}\downarrow$& $\text{MIA}\downarrow$\\
					\midrule
					\multirow{2}{*}{Simple}  &  $100.00$ & $100.00$ & $94.76$ & $96.08$ &  $93.19$\\
                    &  $99.99$ & $0.00$ & $95.25$ & $0.03$ &  $0.00$\\
                    \cmidrule{1-6}
					\multirow{2}{*}{Contrastive} & $99.81$         &$99.54$            &$94.47$            & $93.74$  & $86.82$\\ 
                                                    & $99.94$         &$0.00$            &$94.75$            & $0.00$  & $0.01$\\
                    \cmidrule{1-6}
                    \multirow{2}{*}{Cutout}   & $100.00$         &$100.00$            &$96.02$            & $96.44$  & $89.50$\\
                                              & $99.95$         &$0.06$            &$95.98$            & $0.01$  & $0.00$\\
					\bottomrule
				\end{tabular}	
	}
\end{table}
\section{Conclusion}
\label{sec:5_conclusion}
This study introduces a fresh viewpoint on unlearning, focusing on the diverse reactions of neural networks toward the augmentations of forgetting and remaining data. Through this lens, we introduce the novel concepts of knowledge vaporization and concentration to selectively eliminate the knowledge acquired from the forgetting data. Based on the proposed concepts, we employed Siamese networks to develop an unlearning approach that efficiently erases forgetting data without incurring additional memory costs and only with restricted access to the remaining data. Extensive experimental results demonstrated its efficient forgetting performance in erasing forgetting data, maintaining model utility, and protecting privacy. 
{
    \small
    \bibliographystyle{ieeenat_fullname}
    \bibliography{main}

\begin{thebibliography}{34}
\providecommand{\natexlab}[1]{#1}
\providecommand{\url}[1]{\texttt{#1}}
\expandafter\ifx\csname urlstyle\endcsname\relax
  \providecommand{\doi}[1]{doi: #1}\else
  \providecommand{\doi}{doi: \begingroup \urlstyle{rm}\Url}\fi

\bibitem[Bertinetto et~al.(2016)Bertinetto, Valmadre, Henriques, Vedaldi, and Torr]{bertinetto2016fully}
Luca Bertinetto, Jack Valmadre, Joao~F Henriques, Andrea Vedaldi, and Philip~HS Torr.
\newblock Fully-convolutional siamese networks for object tracking.
\newblock In \emph{Computer Vision--ECCV 2016 Workshops: Amsterdam, The Netherlands, October 8-10 and 15-16, 2016, Proceedings, Part II 14}, pages 850--865. Springer, 2016.

\bibitem[Bourtoule et~al.(2021)Bourtoule, Chandrasekaran, Choquette-Choo, Jia, Travers, Zhang, Lie, and Papernot]{bourtoule2021machine}
Lucas Bourtoule, Varun Chandrasekaran, Christopher~A Choquette-Choo, Hengrui Jia, Adelin Travers, Baiwu Zhang, David Lie, and Nicolas Papernot.
\newblock Machine unlearning.
\newblock In \emph{2021 IEEE Symposium on Security and Privacy (SP)}, pages 141--159. IEEE, 2021.

\bibitem[Bromley et~al.(1993)Bromley, Guyon, LeCun, S{\"a}ckinger, and Shah]{bromley1993signature}
Jane Bromley, Isabelle Guyon, Yann LeCun, Eduard S{\"a}ckinger, and Roopak Shah.
\newblock Signature verification using a" siamese" time delay neural network.
\newblock \emph{Advances in neural information processing systems}, 6, 1993.

\bibitem[Cao and Yang(2015)]{cao2015towards}
Yinzhi Cao and Junfeng Yang.
\newblock Towards making systems forget with machine unlearning.
\newblock In \emph{2015 IEEE symposium on security and privacy}, pages 463--480. IEEE, 2015.

\bibitem[Chen et~al.(2022)Chen, Zhang, Wang, Backes, Humbert, and Zhang]{chen2022graph}
Min Chen, Zhikun Zhang, Tianhao Wang, Michael Backes, Mathias Humbert, and Yang Zhang.
\newblock Graph unlearning.
\newblock In \emph{Proceedings of the 2022 ACM SIGSAC conference on computer and communications security}, pages 499--513, 2022.

\bibitem[Chen et~al.(2023)Chen, Gao, Liu, Peng, and Wang]{chen2023boundary}
Min Chen, Weizhuo Gao, Gaoyang Liu, Kai Peng, and Chen Wang.
\newblock Boundary unlearning: Rapid forgetting of deep networks via shifting the decision boundary.
\newblock In \emph{Proceedings of the IEEE/CVF Conference on Computer Vision and Pattern Recognition}, pages 7766--7775, 2023.

\bibitem[Chen et~al.(2020)Chen, Kornblith, Norouzi, and Hinton]{chen2020simple}
Ting Chen, Simon Kornblith, Mohammad Norouzi, and Geoffrey Hinton.
\newblock A simple framework for contrastive learning of visual representations.
\newblock In \emph{International conference on machine learning}, pages 1597--1607. PMLR, 2020.

\bibitem[Chen and He(2021)]{chen2021exploring}
Xinlei Chen and Kaiming He.
\newblock Exploring simple siamese representation learning.
\newblock In \emph{Proceedings of the IEEE/CVF conference on computer vision and pattern recognition}, pages 15750--15758, 2021.

\bibitem[Chundawat et~al.(2023{\natexlab{a}})Chundawat, Tarun, Mandal, and Kankanhalli]{chundawat2023can}
Vikram~S Chundawat, Ayush~K Tarun, Murari Mandal, and Mohan Kankanhalli.
\newblock Can bad teaching induce forgetting? unlearning in deep networks using an incompetent teacher.
\newblock In \emph{Proceedings of the AAAI Conference on Artificial Intelligence}, pages 7210--7217, 2023{\natexlab{a}}.

\bibitem[Chundawat et~al.(2023{\natexlab{b}})Chundawat, Tarun, Mandal, and Kankanhalli]{chundawat2023zero}
Vikram~S Chundawat, Ayush~K Tarun, Murari Mandal, and Mohan Kankanhalli.
\newblock Zero-shot machine unlearning.
\newblock \emph{IEEE Transactions on Information Forensics and Security}, 18:\penalty0 2345--2354, 2023{\natexlab{b}}.

\bibitem[DeVries and Taylor(2017)]{devries2017improved}
Terrance DeVries and Graham~W Taylor.
\newblock Improved regularization of convolutional neural networks with cutout.
\newblock \emph{arXiv preprint arXiv:1708.04552}, 2017.

\bibitem[Foster et~al.(2024)Foster, Schoepf, and Brintrup]{foster2024fast}
Jack Foster, Stefan Schoepf, and Alexandra Brintrup.
\newblock Fast machine unlearning without retraining through selective synaptic dampening.
\newblock In \emph{Proceedings of the AAAI Conference on Artificial Intelligence}, pages 12043--12051, 2024.

\bibitem[Ginart et~al.(2019)Ginart, Guan, Valiant, and Zou]{ginart2019making}
Antonio Ginart, Melody Guan, Gregory Valiant, and James~Y Zou.
\newblock Making ai forget you: Data deletion in machine learning.
\newblock \emph{Advances in neural information processing systems}, 32, 2019.

\bibitem[Golatkar et~al.(2020)Golatkar, Achille, and Soatto]{golatkar2020eternal}
Aditya Golatkar, Alessandro Achille, and Stefano Soatto.
\newblock Eternal sunshine of the spotless net: Selective forgetting in deep networks.
\newblock In \emph{Proceedings of the IEEE/CVF Conference on Computer Vision and Pattern Recognition}, pages 9304--9312, 2020.

\bibitem[Graves et~al.(2021)Graves, Nagisetty, and Ganesh]{graves2021amnesiac}
Laura Graves, Vineel Nagisetty, and Vijay Ganesh.
\newblock Amnesiac machine learning.
\newblock In \emph{Proceedings of the AAAI Conference on Artificial Intelligence}, pages 11516--11524, 2021.

\bibitem[He et~al.(2016)He, Zhang, Ren, and Sun]{he2016deep}
Kaiming He, Xiangyu Zhang, Shaoqing Ren, and Jian Sun.
\newblock Deep residual learning for image recognition.
\newblock In \emph{Proceedings of the IEEE conference on computer vision and pattern recognition}, pages 770--778, 2016.

\bibitem[Koch et~al.(2015)Koch, Zemel, Salakhutdinov, et~al.]{koch2015siamese}
Gregory Koch, Richard Zemel, Ruslan Salakhutdinov, et~al.
\newblock Siamese neural networks for one-shot image recognition.
\newblock In \emph{ICML deep learning workshop}, pages 1--30. Lille, 2015.

\bibitem[Koh and Liang(2017)]{koh2017understanding}
Pang~Wei Koh and Percy Liang.
\newblock Understanding black-box predictions via influence functions.
\newblock In \emph{International conference on machine learning}, pages 1885--1894. PMLR, 2017.

\bibitem[Krizhevsky et~al.(2009)Krizhevsky, Hinton, et~al.]{krizhevsky2009learning}
Alex Krizhevsky, Geoffrey Hinton, et~al.
\newblock Learning multiple layers of features from tiny images.
\newblock 2009.

\bibitem[Kurmanji et~al.(2024)Kurmanji, Triantafillou, Hayes, and Triantafillou]{kurmanji2024towards}
Meghdad Kurmanji, Peter Triantafillou, Jamie Hayes, and Eleni Triantafillou.
\newblock Towards unbounded machine unlearning.
\newblock \emph{Advances in neural information processing systems}, 36, 2024.

\bibitem[Lin et~al.(2023)Lin, Zhang, Chen, Chen, and Susilo]{lin2023erm}
Shen Lin, Xiaoyu Zhang, Chenyang Chen, Xiaofeng Chen, and Willy Susilo.
\newblock Erm-ktp: Knowledge-level machine unlearning via knowledge transfer.
\newblock In \emph{Proceedings of the IEEE/CVF Conference on Computer Vision and Pattern Recognition}, pages 20147--20155, 2023.

\bibitem[Martens(2020)]{martens2020new}
James Martens.
\newblock New insights and perspectives on the natural gradient method.
\newblock \emph{Journal of Machine Learning Research}, 21\penalty0 (146):\penalty0 1--76, 2020.

\bibitem[Nguyen et~al.(2022)Nguyen, Huynh, Nguyen, Liew, Yin, and Nguyen]{nguyen2022survey}
Thanh~Tam Nguyen, Thanh~Trung Huynh, Phi~Le Nguyen, Alan Wee-Chung Liew, Hongzhi Yin, and Quoc Viet~Hung Nguyen.
\newblock A survey of machine unlearning.
\newblock \emph{arXiv preprint arXiv:2209.02299}, 2022.

\bibitem[Simonyan and Zisserman(2014)]{simonyan2014very}
Karen Simonyan and Andrew Zisserman.
\newblock Very deep convolutional networks for large-scale image recognition.
\newblock \emph{arXiv preprint arXiv:1409.1556}, 2014.

\bibitem[Song and Mittal(2021)]{song2021systematic}
Liwei Song and Prateek Mittal.
\newblock Systematic evaluation of privacy risks of machine learning models.
\newblock In \emph{30th USENIX Security Symposium (USENIX Security 21)}, pages 2615--2632, 2021.

\bibitem[Taigman et~al.(2014)Taigman, Yang, Ranzato, and Wolf]{taigman2014deepface}
Yaniv Taigman, Ming Yang, Marc'Aurelio Ranzato, and Lior Wolf.
\newblock Deepface: Closing the gap to human-level performance in face verification.
\newblock In \emph{Proceedings of the IEEE conference on computer vision and pattern recognition}, pages 1701--1708, 2014.

\bibitem[Tarun et~al.(2024)Tarun, Chundawat, Mandal, and Kankanhalli]{10113700}
Ayush~K. Tarun, Vikram~S. Chundawat, Murari Mandal, and Mohan Kankanhalli.
\newblock Fast yet effective machine unlearning.
\newblock \emph{IEEE Transactions on Neural Networks and Learning Systems}, 35\penalty0 (9):\penalty0 13046--13055, 2024.

\bibitem[Thudi et~al.(2022)Thudi, Jia, Shumailov, and Papernot]{thudi2022necessity}
Anvith Thudi, Hengrui Jia, Ilia Shumailov, and Nicolas Papernot.
\newblock On the necessity of auditable algorithmic definitions for machine unlearning.
\newblock In \emph{31st USENIX Security Symposium (USENIX Security 22)}, pages 4007--4022, 2022.

\bibitem[Tian et~al.(2020)Tian, Sun, Poole, Krishnan, Schmid, and Isola]{tian2020makes}
Yonglong Tian, Chen Sun, Ben Poole, Dilip Krishnan, Cordelia Schmid, and Phillip Isola.
\newblock What makes for good views for contrastive learning?
\newblock \emph{Advances in neural information processing systems}, 33:\penalty0 6827--6839, 2020.

\bibitem[Ullah et~al.(2021)Ullah, Mai, Rao, Rossi, and Arora]{ullah2021machine}
Enayat Ullah, Tung Mai, Anup Rao, Ryan~A Rossi, and Raman Arora.
\newblock Machine unlearning via algorithmic stability.
\newblock In \emph{Conference on Learning Theory}, pages 4126--4142. PMLR, 2021.

\bibitem[Veale and Zuiderveen~Borgesius(2021)]{veale2021demystifying}
Michael Veale and Frederik Zuiderveen~Borgesius.
\newblock Demystifying the draft eu artificial intelligence act—analysing the good, the bad, and the unclear elements of the proposed approach.
\newblock \emph{Computer Law Review International}, 22\penalty0 (4):\penalty0 97--112, 2021.

\bibitem[Voigt and Von~dem Bussche(2017)]{voigt2017eu}
Paul Voigt and Axel Von~dem Bussche.
\newblock The eu general data protection regulation (gdpr).
\newblock \emph{A Practical Guide, 1st Ed., Cham: Springer International Publishing}, 10\penalty0 (3152676):\penalty0 10--5555, 2017.

\bibitem[Warnecke et~al.(2021)Warnecke, Pirch, Wressnegger, and Rieck]{warnecke2021machine}
Alexander Warnecke, Lukas Pirch, Christian Wressnegger, and Konrad Rieck.
\newblock Machine unlearning of features and labels.
\newblock \emph{arXiv preprint arXiv:2108.11577}, 2021.

\bibitem[Warner(1965)]{warner1965randomized}
Stanley~L Warner.
\newblock Randomized response: A survey technique for eliminating evasive answer bias.
\newblock \emph{Journal of the American statistical association}, 60\penalty0 (309):\penalty0 63--69, 1965.

\end{thebibliography}
}

\onecolumn
\setcounter{secnumdepth}{2}
\section{Empirical Analysis on the Retrained Models}
To find an effective approach to approximate retrained models, we conduct experiments to empirically examine the behaviors of retrained models on both forgetting and remaining data. Using the CIFAR-10 dataset, we train the VGG16-BN solely on the remaining data across three unlearning scenarios: 
\begin{enumerate}
    \item \textbf{Full-class unlearning:} The classes \texttt{automobile}, \texttt{cat}, and \texttt{truck} are removed from the CIFAR-10 dataset.
    \item \textbf{Sub-class unlearning:} We focus on the \texttt{automobile} class, randomly selecting $90\%$ of its training data as forgetting data.
    \item \textbf{Random unlearning:} $10\%$ of the training data points are designated as forgetting points and excluded from the dataset.
\end{enumerate}

Moreover, we utilize data augmentation and observe the response of the retrained models toward the augmentations of data points. To comprehensively investigate the behaviors of the retrained models, we adopt three different types of data augmentation:
\begin{itemize}
    \item \textbf{Simple}: The simple augmentation only consists of a random crop and a random horizontal flip with a probability of $50\%$.
    \item \textbf{Contrastive:} The contrastive data augmentation is widely adopted in contrastive learning methods. It consists of the following different augmentations. First, the geometric augmentation consists of a random resized crop with a scale from 0.2 to 1.0 and a random horizontal flip. Color augmentation consists of randomly changing the brightness, contrast, saturation, and hue of an image with a probability of $80\%$ and randomly converting images to grayscale with a probability of $20\%$. Blurring augmentation is applied with a Gaussian kernel of from 0.1 to 2.0 std. 
    \item \textbf{Cutout:} Cutout is a recent data augmentation method where random square regions are removed from training images, forcing the model to learn more robust features that are invariant to local occlusions or corruptions.
\end{itemize}

\begin{figure}
\centering
\begin{subfigure}{0.47\linewidth}
  \centering
  \includegraphics[width=.999\linewidth]{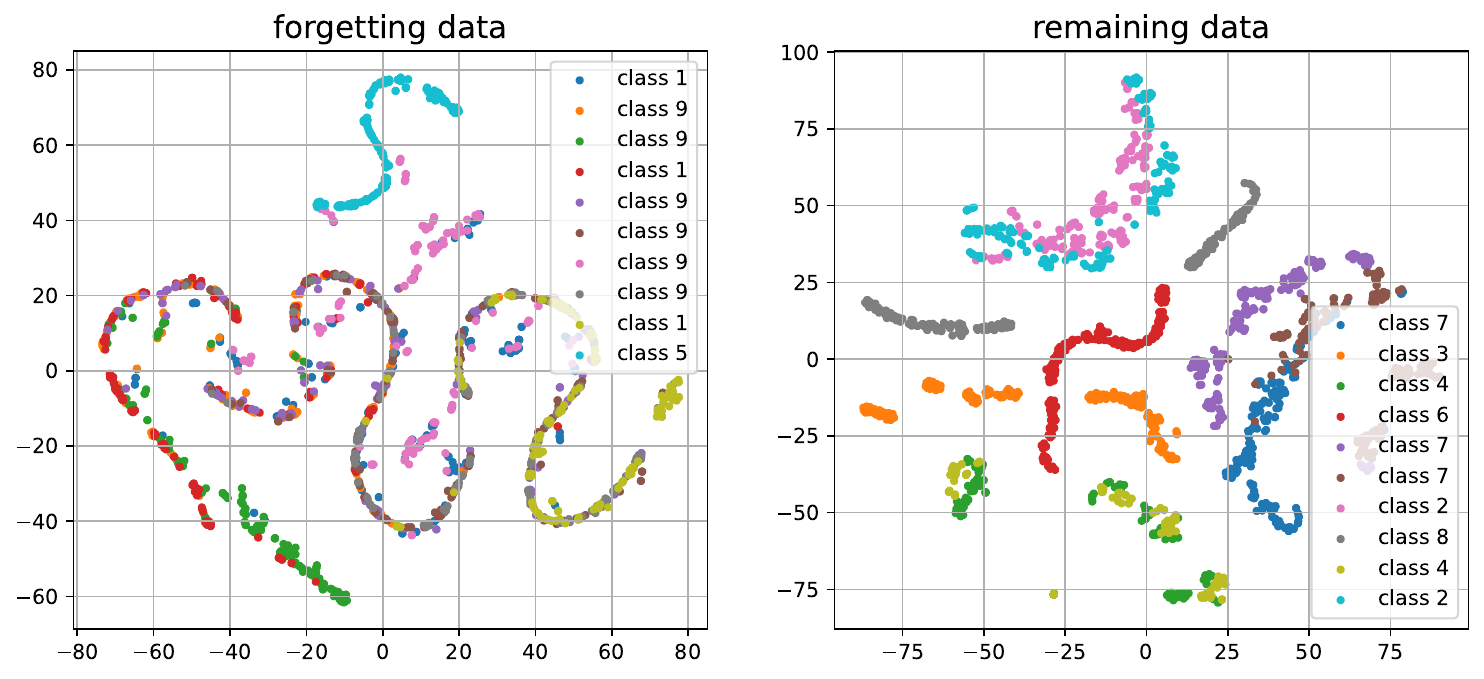}
  \caption{Forgetting full-classes of \texttt{automobile}, \texttt{cat}, and \texttt{truck}}
\end{subfigure}%
\begin{subfigure}{0.47\linewidth}
  \centering
  \includegraphics[width=.999\linewidth]{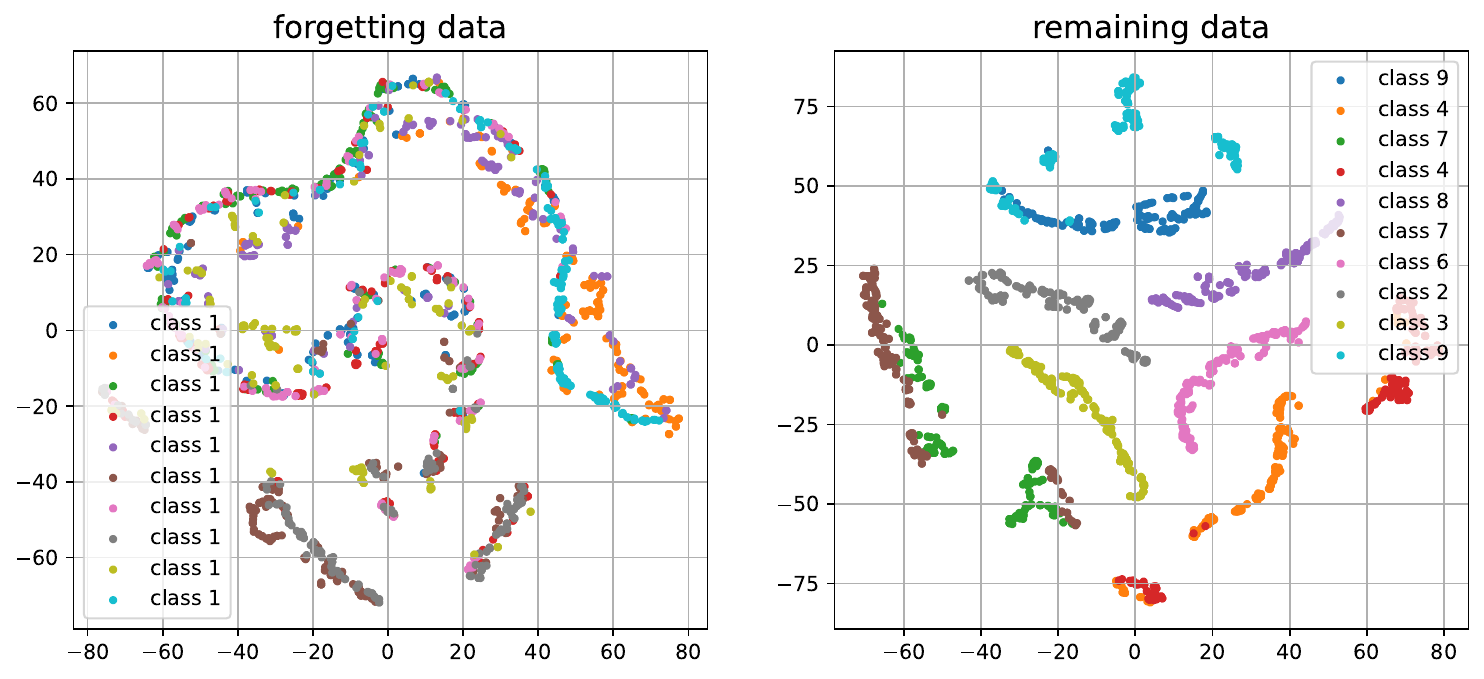}
  \caption{Forgetting 90\% data samples of class \texttt{automobile}}
\end{subfigure}
\begin{subfigure}{0.47\linewidth}
  \centering
  \includegraphics[width=.999\linewidth]{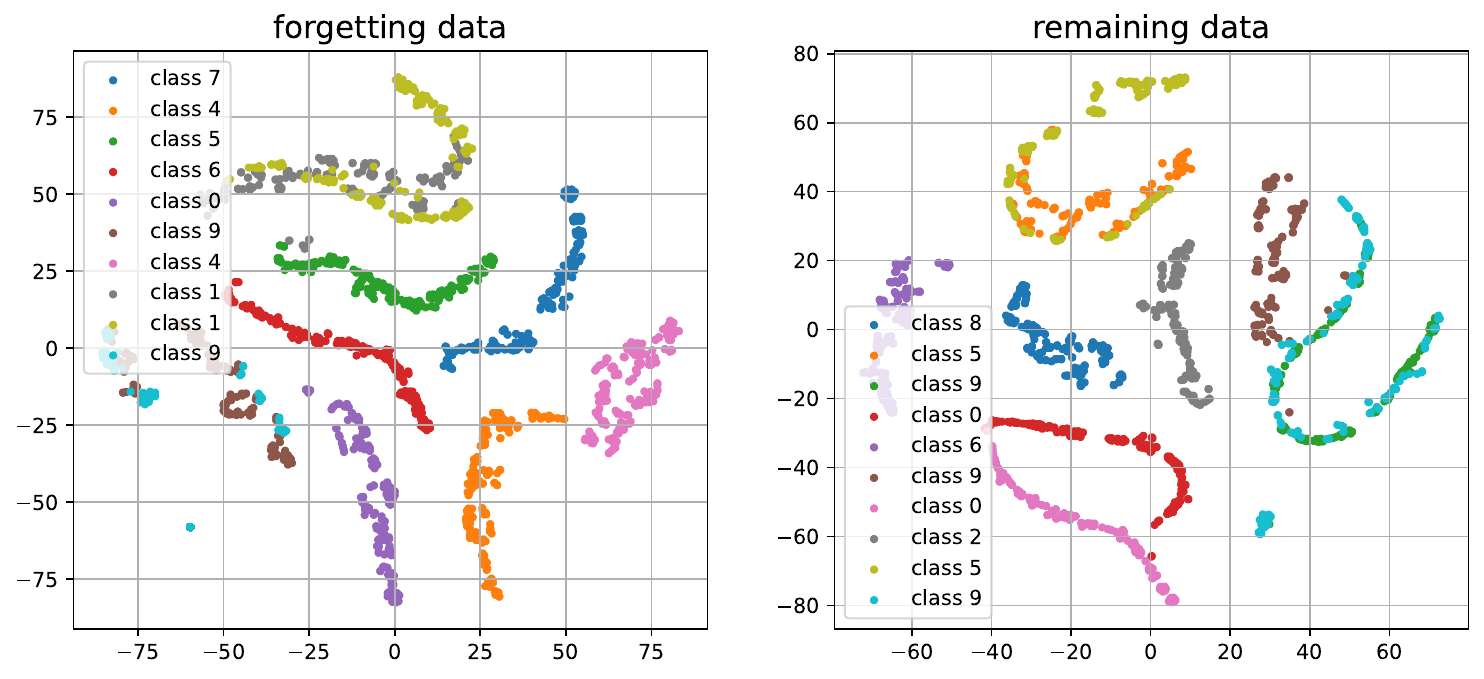}
  \caption{Forgetting 10\% random data samples }
\end{subfigure}
\caption{$t$-SNE visualization of logit outputs from retrained models with the simple data augmentation in three scenarios: (a) full-class forgetting, (b) sub-class forgetting, and (c) random forgetting.} 
\label{app_fig:intro_fig_simple}
\end{figure}

\begin{figure}
\centering
\begin{subfigure}{0.47\linewidth}
  \centering
  \includegraphics[width=.999\linewidth]{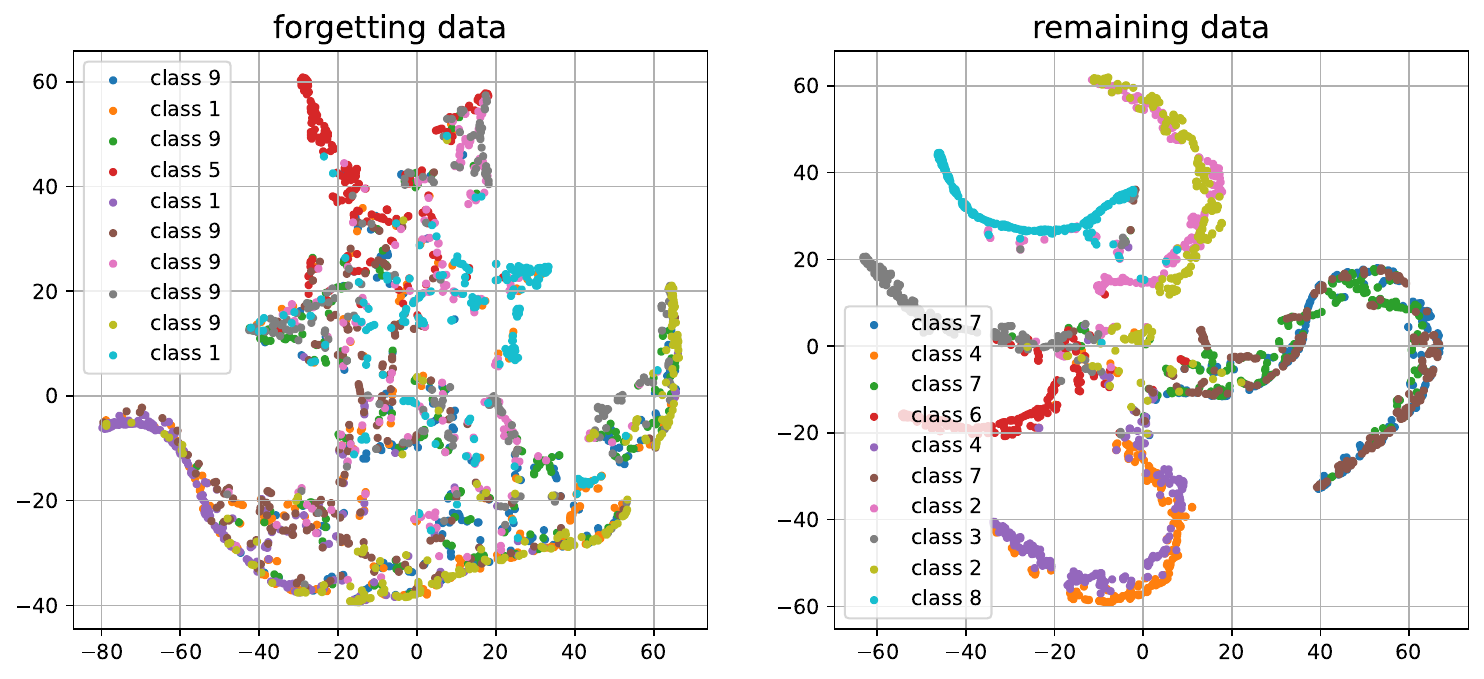}
  \caption{Forgetting full-classes of \texttt{automobile}, \texttt{cat}, and \texttt{truck}}
\end{subfigure}%
\begin{subfigure}{0.47\linewidth}
  \centering
  \includegraphics[width=.999\linewidth]{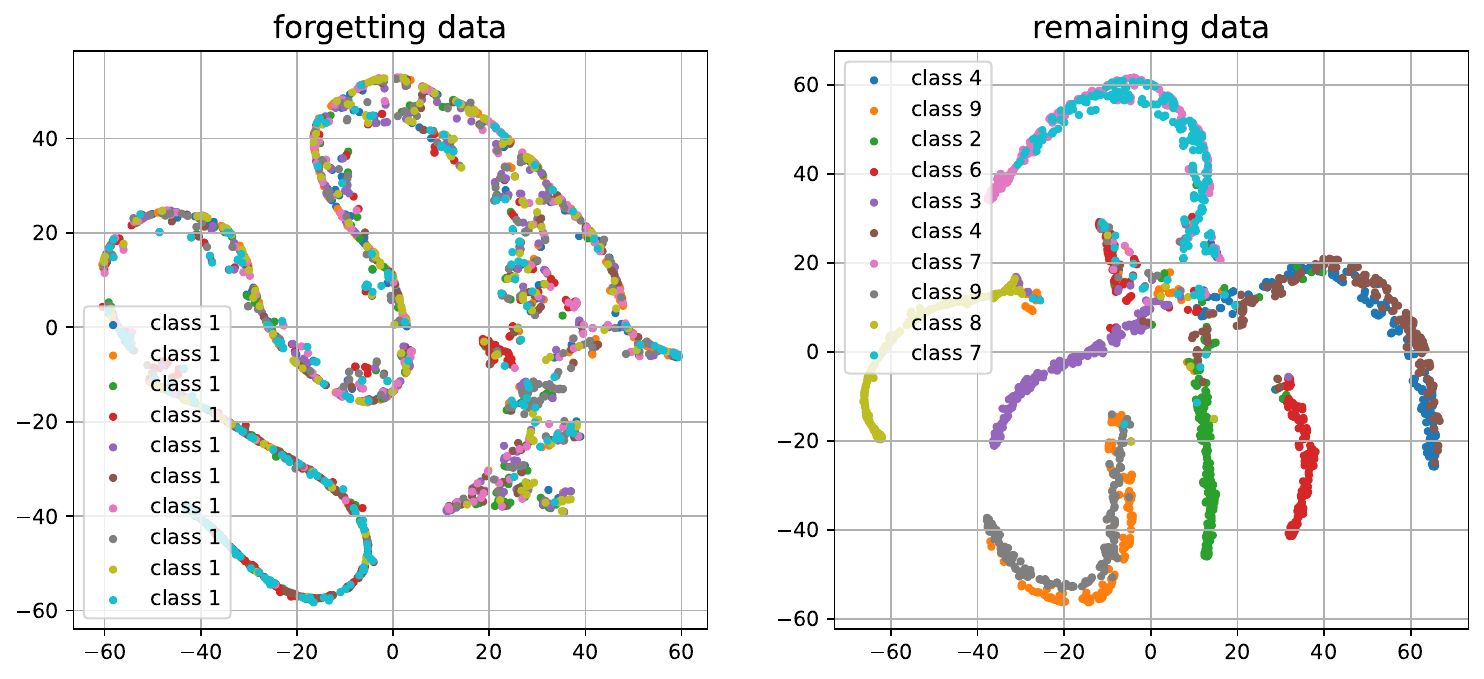}
  \caption{Forgetting 90\% data samples of class \texttt{automobile}}
\end{subfigure}
\begin{subfigure}{0.47\linewidth}
  \centering
  \includegraphics[width=.999\linewidth]{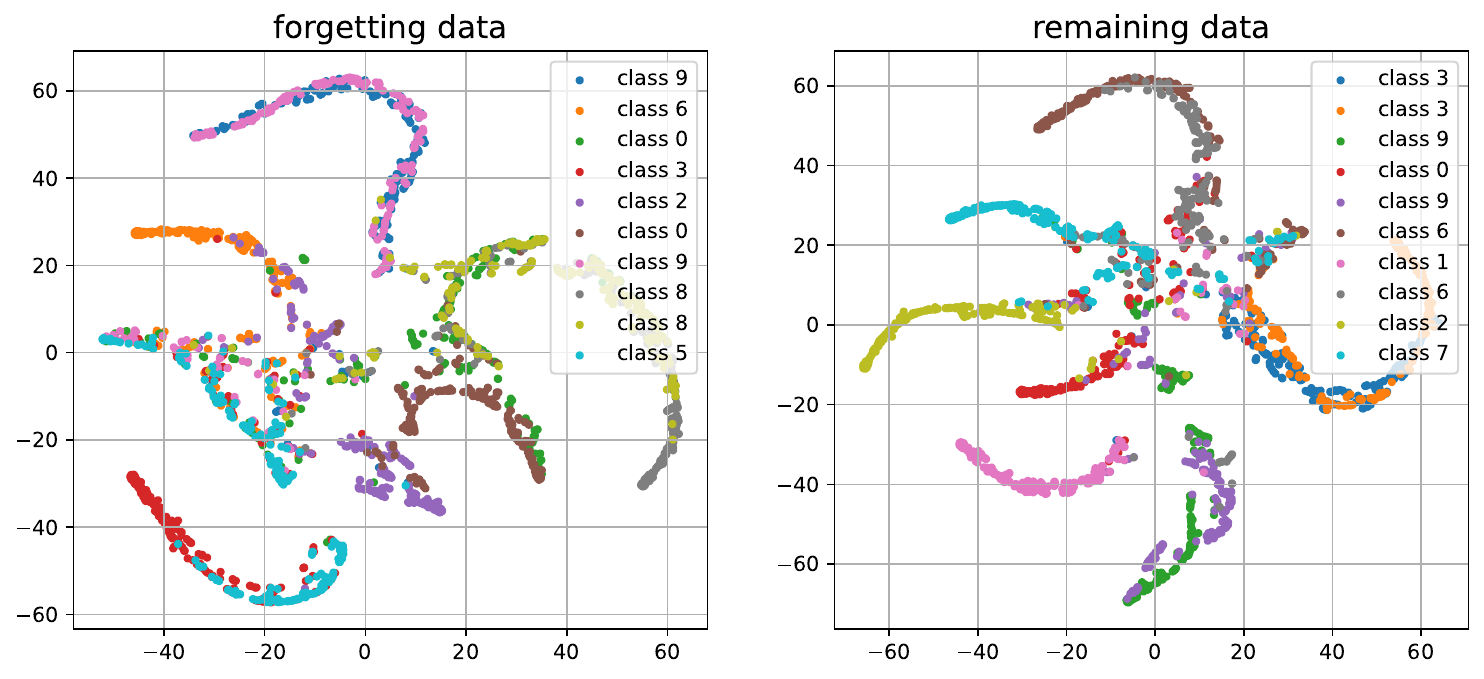}
  \caption{Forgetting 10\% random data samples }
\end{subfigure}
\caption{$t$-SNE visualization of logit outputs from retrained models with the contrastive data augmentation in three scenarios: (a) full-class forgetting, (b) sub-class forgetting, and (c) random forgetting.} 
\label{app_fig:intro_fig_contrastive}
\end{figure}

\begin{figure}
\centering
\begin{subfigure}{0.47\linewidth}
  \centering
  \includegraphics[width=.999\linewidth]{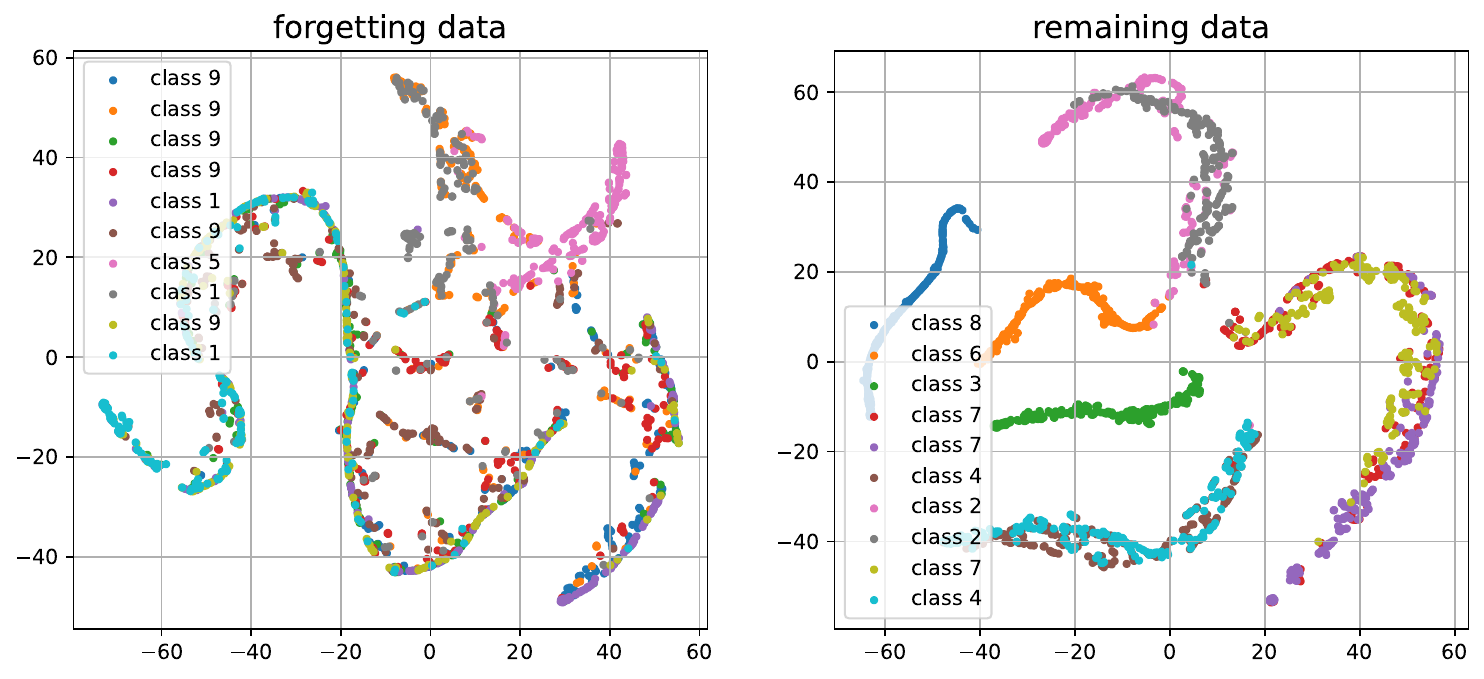}
  \caption{Forgetting full-classes of \texttt{automobile}, \texttt{cat}, and \texttt{truck}}
\end{subfigure}%
\begin{subfigure}{0.47\linewidth}
  \centering
  \includegraphics[width=.999\linewidth]{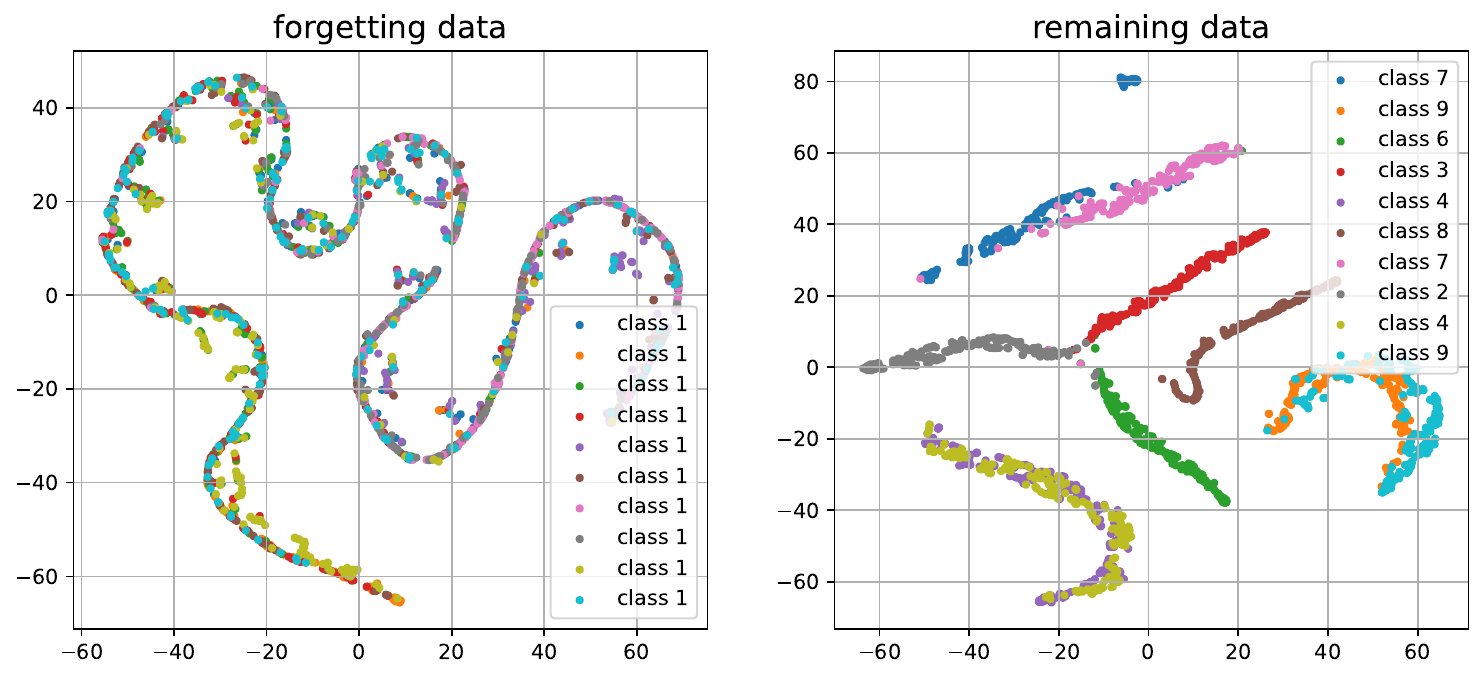}
  \caption{Forgetting 90\% data samples of class \texttt{automobile}}
\end{subfigure}
\begin{subfigure}{0.47\linewidth}
  \centering
  \includegraphics[width=.999\linewidth]{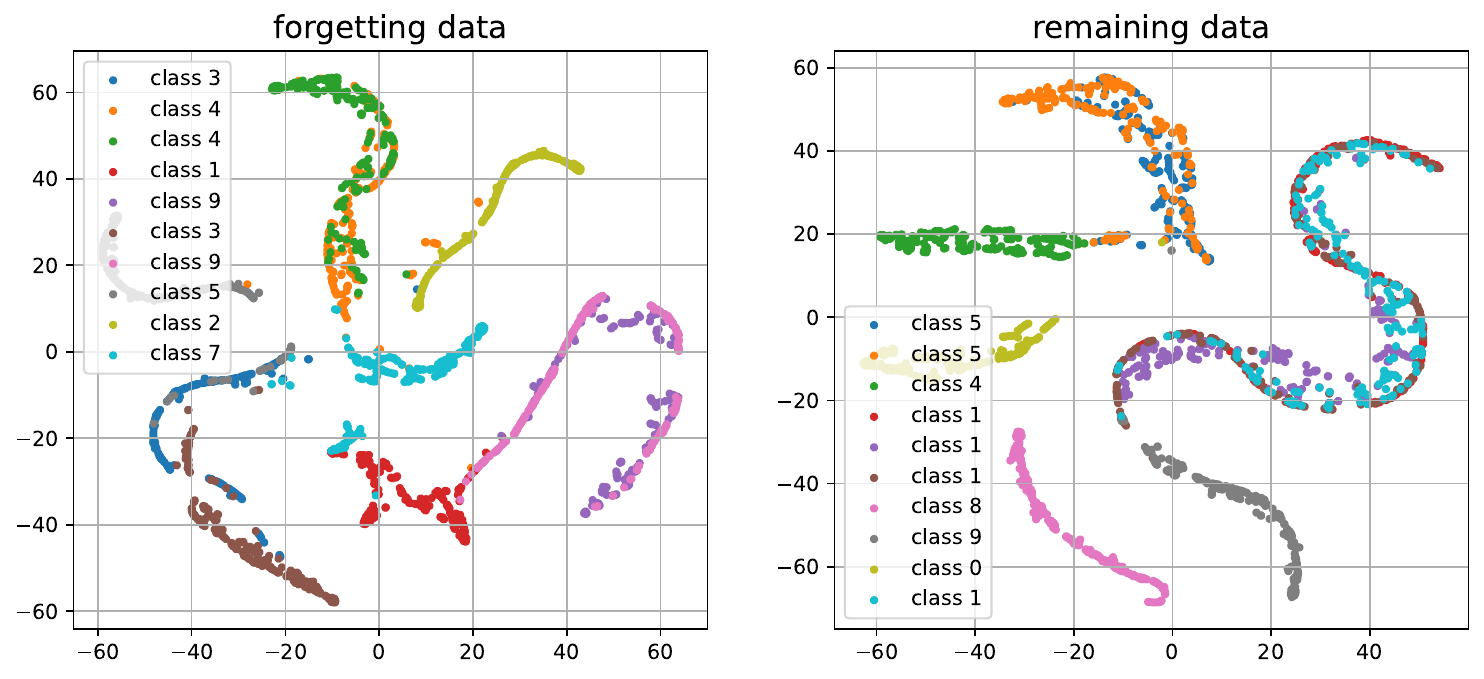}
  \caption{Forgetting 10\% random data samples }
\end{subfigure}
\caption{$t$-SNE visualization of logit outputs from retrained models with the cutout data augmentation in three scenarios: (a) full-class forgetting, (b) sub-class forgetting, and (c) random forgetting.} 
\label{app_fig:intro_fig_random}
\end{figure}

We retrain the VGG16-BN model from scratch on the remaining dataset using the aforementioned data augmentation methods. In each unlearning scenario, we randomly select 10 data points from both the remaining and forgetting datasets, generating 100 augmented views for each data point. Subsequently, we extract the output logits of these augmented views and visualize them in 2-dimensional embeddings via $t$-SNE. The visualization outcomes for the three unlearning scenarios can be observed in Figure~\ref{app_fig:intro_fig_simple}, Figure~\ref{app_fig:intro_fig_contrastive}, and Figure~\ref{app_fig:intro_fig_random}.

For the remaining data, a distinct pattern emerges where the augmented views of the same data (indicated by dots of the same color) are tightly clustered, forming discernible structures. Additionally, the augmented views of data points sharing the same label tend to group together. Conversely, when examining the forgetting data, the augmented views of identical data points exhibit greater dispersion, lacking clear patterns. This observation is particularly notable in full-class and sub-class forgetting scenarios. Notably, the dispersion effect in random unlearning is less pronounced compared to other unlearning scenarios. This discrepancy is attributed to the generalization of the remaining data, which aids in preserving utility for the forgetting data, thereby reducing dispersion in the augmented views.
This observation underscores a crucial aspect of unlearning. It emphasizes that unlearning does not equate to the complete eradication of the model's effectiveness on the forgetting data. Instead, the model retains a certain degree of utility on the forgetting data from a specific class due to the generalization of the remaining data belonging to the same class.

To corroborate our observations, we compute conditional distributions by applying softmax to the logits output from the augmented views of both forgetting and remaining data. Subsequently, we calculate the Kullback-Leibler (KL) divergence for the augmentations of the same data points. The numerical results presented in Table~\ref{app_tab:KL_divergence} support our visual findings, indicating higher KL divergence values for the forgetting data across all three unlearning scenarios. This is logical as data from which no knowledge is retained tends to exhibit greater randomness in the model's representations compared to the remaining data.
\begin{table}
	\centering
    \caption{The averaged KL divergence for the augmented views of the forgetting and remaining data for the VGG16-BN model on the CIFAR-10 dataset under three unlearning scenarios}
	\label{app_tab:KL_divergence}
				\begin{tabular}{lccc}
					\toprule
					Model & Aug & $D_{KL}$ for $\mathcal{D}_r$& $D_{KL}$ for $\mathcal{D}_f$\\
					\midrule
					\multirow{3}{*}{Full-class}  & Simple & $0.0000$ & $1.0782$\\
                                               & Contrastive& $0.8906$ & $1.3737$\\
                                               & Cutout  & $0.2100$ & $1.4675$\\
                    \cmidrule{2-4}
                    \multirow{3}{*}{Sub-class}  & Simple & $0.0004$ & $0.0141$\\
                                               & Contrastive& $1.3192$ & $2.2822$\\
                                               & Cutout  & $0.1716$ & $0.6215$\\
                    \cmidrule{2-4}
                    \multirow{3}{*}{Random}  & Simple & $0.0021$ & $0.1048$\\
                                               & Contrastive& $1.6926$ & $1.7256$\\
                                               & Cutout  & $0.2889$ & $0.7473$\\
					\bottomrule
				\end{tabular}	
\end{table}
\section{Implementation Details}
The unlearning procedures of the proposed Siamese unlearning method are summarized in Algorithm~\ref{algo:siam_unlearning}.
\begin{algorithm}[t]
\caption{Siamese Machine Unlearning}
\begin{algorithmic}[1]
\label{algo:siam_unlearning}
\REQUIRE $T$ (number of epochs), unlearned data $\mathcal{D}_f$, few remaining data samples $\mathcal{S}_f$, the hyperparameter $\lambda$, batch size for the forgetting data $M_f$, and batch size for the remaining data $M_r$.
\STATE Compute the ratios of unlearned data within each class, $\boldsymbol{r} = (r_1, r_2, \dots, r_K)$.
\STATE Establish the adaptive permutation $p$ according to $\boldsymbol{r}$.
\STATE $\boldsymbol{\omega} \leftarrow \boldsymbol{\omega}_0$
\WHILE{iteration $t=1$ to $T$}
        \STATE Get a mini-batch data $\{\x_m, y_m \}_{m=1}^{M_r}$ from $\mathcal{D}_r$
        \STATE Get a mini-batch data $\{\x_m, y_m \}_{m=1}^{M_f}$ from $\mathcal{D}_f$
        \STATE Compute the knowledge concentration loss $\mathcal{L}_{KC}(\boldsymbol{\omega}, \x) + \lambda \text{SCE}(\x, y)$ for $\{\x_m, y_m \}_{m=1}^{M_r}$
        \STATE Update the parameters $\boldsymbol{\omega}$ through backpropagation.
        \STATE Compute the knowledge vaporization loss $\mathcal{L}_{KV}(\boldsymbol{\omega}, \x) + \lambda \text{SCE}(\x, p(y))$ for $\{\x_m, y_m \}_{m=1}^{M_f}$
        \STATE Update the parameters $\boldsymbol{\omega}$ through backpropagation.
\ENDWHILE
\end{algorithmic}
\end{algorithm}
\paragraph{Predictor design.}
The predictor $h$ in the proposed Siamese unlearning method comprises an MLP head consisting of two fully connected layers with a batch normalization layer. Given that $h$ simply transforms the output logits $\boldsymbol{l} \in \mathbb{R}^{10}$, we opt for a hidden layer size of only 64. This size is significantly smaller in comparison to the dimensions of the entire model.

\section{Experiment Details}
\subsection{Training Settings}
\paragraph{Pretraining} 
In the experiments, pretraining of the original and retrained models is necessary to evaluate the performance of unlearning. For the three unlearning scenarios, both the original and retrained models on the CIFAR-10 dataset adopt the VGG16-BN architecture. Similarly, for the CIFAR-100 dataset, the architectures used are ResNet18 and ResNet50 for the original and retrained models. All models undergo pretraining for 200 epochs with a warm-up period of 2 epochs. An SGD optimizer with a learning rate of $0.001$ and weight decay of $0.0001$ for all model parameters is employed. Additionally, a step learning rate scheduler is utilized, with the learning rate decaying at the 60th, 120th, and 160th epochs. To mirror practical scenarios, the three data augmentations detailed in Appendix A are applied during pretraining. Training is conducted on a server equipped with an NVIDIA A40 GPU.

\paragraph{Unlearning.} 
To reproduce the compared methods, we re-implement them based on each paper's specifications and official public source codes. When tuning hyperparameters for the compared methods, we prioritize achieving optimal metric values across different scenarios. Notably, all compared baselines utilize the entire remaining dataset $\mathcal{D}_r$ for unlearning, whereas the proposed Siamese unlearning methods only utilize a limited subset of the remaining data $\mathcal{S}_r$ consisting of 1000 samples. It is crucial to highlight that the size of $\mathcal{S}_r$ is less than $2\%$ of the total dataset size for CIFAR-10 and CIFAR-100. For the optimizer in our unlearning method, an SGD optimizer with a learning rate of $0.0001$, momentum of $0.9$, and weight decay of $0.0001$ is employed. Given the limited epochs on the forgetting data $\mathcal{D}_f$, a learning rate scheduler is not utilized for the optimizer during unlearning procedures.
\subsection{Evaluation Metrics}
\paragraph{Acc and TA.} 
To assess the unlearning performance of the evaluated methods, we initially calculate the prediction accuracy (training accuracy) of the unlearned models on the forgetting data $\mathcal{D}_f$ and the remaining data $\mathcal{D}_r$. Specifically, for a dataset $\mathcal{D}$ with a size of $N$, the top-1 accuracy is defined as:
\begin{align}
    \text{Acc}_{\mathcal{D}} = \frac{1}{N} \sum\limits_{(\x, y) \in \mathcal{D}} \mathbb{I}[\arg\max(f(\x) = y)].
\end{align}
It is important to note that $\text{Acc}_{\mathcal{D}_r}$ and $\text{Acc}_{\mathcal{D}_f}$ solely rely on the training set. Regarding performance on the test set, different considerations are made for each unlearning scenario. In the random forgetting scenario, only the test accuracy on the entire test set is evaluated. For full-class and sub-class unlearning scenarios, the test accuracy on samples from the forgetting and remaining classes is taken into account, denoted as $\text{TA}_{\mathcal{D}_f}$ and $\text{TA}_{\mathcal{D}_r}$, respectively.

\paragraph{MIA.} 
Membership inference attack (MIA) aims to determine whether a specific data point was part of the training dataset for a particular model. Consequently, a lower MIA rate on the forgetting data signifies better performance of the unlearned models. Widely utilized in machine unlearning literature, MIA offers an alternative perspective for evaluating the efficacy of unlearning methods in eliminating undesired knowledge from the forgetting data. Various methods exist for conducting MIA, and to maintain consistency with existing unlearning literature, we employ an entropy-based approach where the MIA predictor infers membership by examining the entropy of the output logits. Logistic regression is utilized as the classifier in the MIA process.

\subsection{More Experimental Results of Ablation Studies}
Here we present further experimental results from ablation studies conducted under various unlearning scenarios.
\begin{table}
	\centering
    \caption{Ablation study on CIFAR-10 under full-class unlearning scenario. KVC denotes the loss terms of knowledge vaporization and concentration, SCE denotes the symmetric cross-entropy term, and ALP denote the adaptive label permutation.}
	\label{app-tab:ablation-alp-1}
	\resizebox{0.6\linewidth}{!}{
				\begin{tabular}{lllccccc}
					\toprule
					KVC & CE &ALP &$\text{Acc}_{\mathcal{D}_r} \uparrow$ & $\text{Acc}_{\mathcal{D}_f}\downarrow$ & $\text{TA}_{\mathcal{D}_r}\uparrow$& $\text{TA}_{\mathcal{D}_f}\downarrow$& $\text{MIA}\downarrow$\\
					\midrule
                    & \CheckmarkBold &  & $99.86$ & $\mathbf{0.00}$ & $93.58$ & $\mathbf{0.00}$ & $\underline{0.03}$\\
                    & \CheckmarkBold & \CheckmarkBold &$\underline{99.96}$ &$\mathbf{0.00}$ & $93.63$&  $\mathbf{0.00}$ & $\mathbf{0.00}$ \\
                    \CheckmarkBold &  &  & $99.95$ & $100.00$ & $92.40$ & $96.09$ & $99.83$ \\
                    \CheckmarkBold & \CheckmarkBold &  & $\mathbf{99.98}$ & $\mathbf{0.00}$ & $\underline{94.08}$ & $\mathbf{0.00}$ & $\mathbf{0.00}$ \\
                    \CheckmarkBold & \CheckmarkBold & \CheckmarkBold & $\mathbf{99.98}$ & $\underline{0.01}$ & $\mathbf{94.29}$ & $\mathbf{0.00}$ &  $\mathbf{0.00}$\\
					\bottomrule
				\end{tabular}	
	}
\end{table}

\begin{table}
	\centering
    \caption{Ablation study on CIFAR-10 under random unlearning scenario. KVC denotes the loss terms of knowledge vaporization and concentration, SCE denotes the symmetric cross-entropy term, and ALP denote the adaptive label permutation.}
	\label{app-tab:ablation-alp-2}
	\resizebox{0.55\linewidth}{!}{
				\begin{tabular}{lllcccc}
		\toprule
		KVC & CE &ALP &$\text{Acc}_{\mathcal{D}_r} \uparrow$ & $\text{Acc}_{\mathcal{D}_f}$ & $\text{TA}_{\mathcal{D}}\uparrow$& $\text{MIA}\downarrow$\\
	\midrule
    & \CheckmarkBold &  & $83.22$ & $82.15_{(-16.99)}$ & $77.15$ & $\mathbf{59.32}$ \\
    & \CheckmarkBold & \CheckmarkBold & $98.45$ & $98.24_{(-0.9)}$ & $91.54$ & $71.18$\\
    \CheckmarkBold &  &  & $98.38$ & $\underline{98.46_{(-0.68)}}$ & $91.19$ & $88.36$ \\
    \CheckmarkBold & \CheckmarkBold & & $\underline{99.20}$ & $\mathbf{99.10_{(-0.04)}}$ & $\underline{92.11}$ & $81.50$ \\
    \CheckmarkBold & \CheckmarkBold & \CheckmarkBold & $\mathbf{99.98}$ & $100.00_{(0.86)}$ & $\mathbf{93.97}$ & $\underline{62.10}$\\
	\bottomrule
				\end{tabular}	
	}
\end{table}

The ablation study of Knowledge Vaporization and Concentration (KVC), Cross-Entropy (CE), and Adaptive Label Permutation (ALP) under full-class unlearning scenarios is detailed in Table~\ref{app-tab:ablation-alp-1}. Without cross-entropy terms, optimizing knowledge vaporization and concentration alone fails to alter the output logits of the forgetting data. This can be attributed to the optimization focusing solely on maximizing the distance between the predictor $h$ and $\boldsymbol{p}1 = h(f{\boldsymbol{\omega}}(\boldsymbol{x}1))$, neglecting the impact on $f{\boldsymbol{\omega}}$. Conversely, optimizing solely on the Siamese network with cross-entropy terms, excluding knowledge vaporization and concentration, is reminiscent of the RandLab method and leads to a degradation in model performance on the remaining dataset. The adaptive label permutation proves effective in attenuating excessive damage to preserved utility resulting from generalization on the remaining data.

Moreover, the ablation study under random unlearning scenarios is outlined in Table~\ref{app-tab:ablation-alp-2}. In this table, our proposed method demonstrates superior performance across $\text{Acc}_{\mathcal{D}r}$, $\text{Acc}{\mathcal{D}_f}$, and $\text{MIA}$. While optimizing solely on cross-entropy loss achieves the lowest MIA score, it significantly compromises the accuracy of the unlearned model on both remaining and forgetting data. The comprehensive ablation study underscores the critical importance of the introduced cross-entropy terms and the proposed adaptive label permutation within the framework of knowledge vaporization and concentration.

\end{document}